\documentclass[11pt]{article}

\usepackage[T1]{fontenc}
\usepackage[utf8]{inputenc}
\usepackage{lmodern}
\usepackage[a4paper,margin=2.5cm]{geometry}
\usepackage{microtype}
\usepackage{amsmath}
\usepackage{graphicx}
\usepackage{booktabs}
\usepackage{caption}
\usepackage{authblk}
\usepackage{newunicodechar}
\usepackage{xurl}
\usepackage[hidelinks]{hyperref}

\newunicodechar{⁴}{\textsuperscript{4}}
\newunicodechar{⁵}{\textsuperscript{5}}
\newunicodechar{⁷}{\textsuperscript{7}}
\newunicodechar{⁹}{\textsuperscript{9}}
\newunicodechar{⁻}{\textsuperscript{\textminus}}
\newunicodechar{−}{\ensuremath{-}}
\newunicodechar{≈}{\ensuremath{\approx}}
\newunicodechar{≤}{\ensuremath{\leq}}
\newunicodechar{≥}{\ensuremath{\geq}}
\newunicodechar{Δ}{\ensuremath{\Delta}}
\newunicodechar{α}{\ensuremath{\alpha}}
\newunicodechar{η}{\ensuremath{\eta}}
\newunicodechar{ρ}{\ensuremath{\rho}}

\newenvironment{references}
  {\section*{References}\small
   \begin{list}{}{\setlength{\leftmargin}{1.5em}\setlength{\itemindent}{-1.5em}%
                  \setlength{\itemsep}{3pt}\setlength{\parsep}{0pt}}\raggedright}
  {\end{list}}

\newcommand{\papertitle}{Machine--human agreement in marking undergraduate assignments is specific to model, prompt and institution}

\title{\papertitle}
\author[a,1,*]{Giulio Corsi}
\author[b,1]{Maryam Abo-Tabik}
\author[c]{Yael Benn}
\author[d]{Georgiana Thorpe Apreutesei}
\author[e,2]{Roni Tibon}
\author[f,2]{Alexandru Marcoci}
\author[d,2]{Deborah Talmi}
\affil[a]{Centre for the Future of Intelligence, University of Cambridge, Cambridge, UK}
\affil[b]{Department of Computer Science, University of Lancashire, Preston, UK}
\affil[c]{School of Psychology, Manchester Metropolitan University, Manchester, UK}
\affil[d]{Department of Psychology, University of Cambridge, Cambridge, UK}
\affil[e]{School of Psychology, University of Nottingham, Nottingham, UK}
\affil[f]{Centre for the Study of Existential Risk, University of Cambridge, Cambridge, UK}
\date{}

\begin{document}

\maketitle

{\let\thefootnote\relax
\footnotetext{* Corresponding author: Giulio Corsi, Centre for the Future of Intelligence, University of Cambridge, Cambridge, UK. Email: \href{mailto:gc540@cam.ac.uk}{gc540@cam.ac.uk}}
\footnotetext{\textsuperscript{1} Joint first authors.\quad \textsuperscript{2} Joint senior authors.}}

\begin{abstract}
Large language models (LLMs) promise to automate one of higher education's most resource-intensive activities: assessing student writing. Existing evaluations are typically limited to a single model, a single prompt and a single institution. Using 761 undergraduate psychology assignments from three UK universities, we evaluated three frontier LLMs, individually and in ensemble, on accuracy, prompt sensitivity, error independence, generalisability and evidential basis. Agreement between human and machine marks varied widely, with no model or prompt optimal across institutions. All models compressed marks towards a common central tendency, overvaluing the weakest assignments and undervaluing the strongest. Because this error was shared, ensembling failed to remove it. Machine marks also rewarded linguistic features more strongly than human marks. Since human--machine agreement was a joint product of model, prompt and institution, we conclude that current LLMs lack evaluative capacity transferable across contexts, and that single-context validation overstates readiness for deployment.


\smallskip
\noindent\textbf{Code:} \url{https://github.com/giuliocorsi/machine-human-marking-agreement}

\noindent\textbf{Replication data:} \url{https://zenodo.org/records/21676733}
\end{abstract}

\section{Introduction}

Generative artificial intelligence (AI) has entered higher education at a pace with few precedents in the history of educational technology. Within a few years of the public release of large language models (LLMs), national surveys indicated that 92\% of UK undergraduate students were using generative AI, and that 88\% had used it in preparing assessed work (Freeman, 2025). Students routinely ask LLMs to critique drafts and estimate the quality of their essays, including in high-stakes settings, while academic staff have begun to experiment with the same systems for marking and feedback (Attewell, 2025a, 2025b). Universities therefore face a consequential decision. Assessment is where learning, credentialing and institutional trust converge, and it consumes a substantial share of academic labour. Delegating any part of it to AI systems promises efficiency and pedagogical opportunity, but risks undermining the core of the relationship between students and universities (Razzaq et al., 2025). Whether current AI systems mark student work dependably enough to warrant such delegation is an empirical question -- but not, we argue, one that agreement with human marks in any single setting can settle, because agreement observed under a single model, prompt and institution may not survive variations in any of these dimensions.

We collected an authentic corpus of 761 undergraduate psychology assignments from three diverse UK universities (the University of Cambridge, Manchester Metropolitan University, and the University of Nottingham), each formally marked under the standard assessment process for the degree. Three frontier models from distinct providers (GPT-5.4, Claude Opus 4.6, Gemini 3 Flash) scored the corpus zero-shot, with prompts drawn from a fully crossed 3 × 3 × 3 factorial design and outputs combined through four aggregation strategies. This design allows us to examine not only how closely machine marks approximate human ones, but also the dependence of model evaluations on the method of elicitation, whether the models' errors are sufficiently independent for aggregation to correct them, whether performance transfers across institutions, and whether machine marks are grounded in the properties human markers are rewarding.

\section{Background}

\subsection{LLM-based essay scoring}

Automated essay scoring (AES) has long been explored as an alternative or complement to manual marking, from early pattern-matching approaches through to neural methods (Attali \& Burstein, 2006; Ke \& Ng, 2019; Taghipour \& Ng, 2016), and the emergence of LLMs has reopened the question of whether machines can meaningfully evaluate open-ended academic writing. Early evidence suggests that LLMs approximate human marks on constrained writing tasks, though rarely at levels sufficient for operational use. Zero-shot GPT scoring of the TOEFL11 corpus reached a quadratic weighted kappa of only .39, rising to .62 when supplemented with hand-engineered linguistic features (Mizumoto \& Eguchi, 2023), and on short second-language essays GPT-4 approached human-rater agreement only when supplied with calibration examples, still falling short of a conventional supervised baseline (Yancey et al., 2023). Subsequent studies have extended these findings across models, languages and assessment settings (Impey et al., 2024; Kim, 2025; W. Li \& Liu, 2024; Pack et al., 2024; Xia et al., 2024; Yavuz et al., 2025), but syntheses indicate that agreement is far from uniform, with the strongest results concentrated in standardised public corpora and markedly weaker performance on smaller, authentic classroom tasks (Favero et al., 2026; H. Li et al., 2025). Evidence from higher education points the same way. Agreement with instructors\textquotesingle{} marks can fail to reach significance at all (Gaggioli et al., 2025), and where machine and human marks diverge, the divergence can be systematic: machine marks have clustered towards the middle of the scale (Lundgren, 2024) and, on undergraduate psychology essays, were more lenient than instructors\textquotesingle{} for weaker work and harsher for stronger work (Wetzler et al., 2025).

\subsection{Prompt sensitivity, evaluator bias and error independence}

Machine scores have also proved unstable when their method of elicitation was varied. Changes to elements such as rubric framing and scoring anchors can shift outputs substantially, including when prompt variations carry no meaning for human readers (Mansour et al., 2024; Sclar et al., 2024; Stureborg et al., 2024), and identical inputs yielded divergent scores across repeated runs of the same model (Pack et al., 2024). The adjacent LLM-as-judge literature documents a family of systematic evaluative biases, including position and verbosity effects (Wang et al., 2024; Zheng et al., 2023) and skewed rating distributions that cluster scores within a narrow band of the scale (Stureborg et al., 2024), alongside emerging evidence that machine scores reward stylistic properties over substance (Wu \& Aji, 2023) and vary systematically with demographic characteristics -- gender, race, and first-language background -- that models can infer from the text itself, with biases most pronounced precisely when that inference is correct (Kwako \& Ormerod, 2024; Yang et al., 2025). Most of this evidence derives from natural language generation evaluation, where miscalibration is a methodological inconvenience rather than a consequence suffered by students, and although the effects of prompt formatting are known to correlate only weakly between models (Sclar et al., 2024), whether marking configurations that suit one model transfer to another, or across institutions, has not, to our knowledge, been examined for student assessment. In principle, this sensitivity is remediable: if models err idiosyncratically, aggregating a panel of them should average the noise away, as second marking does for human markers. Whether their errors are in fact independent has never been tested on authentic student writing.

\subsection{Generalisability and evidential basis}

Least developed is the question of what machine agreement, where it occurs, actually tells us. Cross-institutional generalisability remains largely untested because most AES research evaluates performance within a single course. Whether LLM-based evaluative capacity transfers across the grading cultures and mark distributions that characterise different institutions is, to our knowledge, unexamined. The evidential basis of machine marks has likewise received little direct attention, despite a construct-validity literature predating LLMs, which established that automated systems could match human marks while rewarding essay length and lexical sophistication rather than the constructs assessed (Perelman, 2014; Powers et al., 2002; Williamson et al., 1999). Comparisons with teacher ratings suggest that LLMs weight language-related criteria more heavily than content (Seßler et al., 2025). The distinction is consequential because agreement is a population-level property while evidence use is not: two assessors can converge on a mark while weighing entirely different evidence.

\subsection{The human benchmark}

At the same time, any evaluation of machine marking must also confront the standard it is measured against. In UK psychology degree marking, the discipline of our corpus, correlations between first and second markers ranged from 0.47 to 0.93 across compulsory units and were lower, sometimes non-significant, for optional units (Dracup, 1997), while blind paired marking of university examination answers has yielded inter-marker correlations of 0.47--0.72 (Laming, 1990), and reviews of marking reliability in higher education report coefficients spanning essentially the full range depending on task, discipline and cohort size (Meadows \& Billington, 2005; Newstead \& Dennis, 1994). Assessment criteria are, moreover, interpreted and applied differently even by experienced assessors within the same department (Bloxham et al., 2016). These figures set both a benchmark and a ceiling for the analyses that follow.

\section{Methods}

\subsection{Corpus, ethics and data splits}

The corpus comprised 934 marked undergraduate psychology assignments from three UK universities: the University of Cambridge (n = 248), Manchester Metropolitan University (n = 461), and the University of Nottingham (n = 225). All were summative coursework or examination submissions marked by academic staff as part of routine assessment for the degree, against each institution's published criteria and on the standard UK percentage scale (0--100), on which only fixed points are permitted (17 permitted marks at Cambridge and Nottingham, 32 at Manchester Met). We excluded 172 short-form items a priori as a different assessment genre from assignment-length work, and one assignment for which a model returned no mark owing to an API error, leaving a complete-case corpus of 761 assignments (Cambridge n = 133, Manchester Met n = 456, Nottingham n = 172) spanning 50 modules and 87 distinct assignments. Mark distributions differed markedly across institutions (Cambridge M = 68.0, s.d. = 8.6; Manchester Met M = 62.5, s.d. = 11.2; Nottingham M = 62.6, s.d. = 6.8).

The study received approval from the University of Cambridge Higher Education Studies Research Ethics Committee (CHESREC; reference 2025.LT.84), and students gave informed consent for their submitted work to be used in the research. Submissions were de-identified before processing, and no student-identifying information was supplied to any model.

The corpus was partitioned once, pseudo-randomly within institution, into a 20\% calibration set (n = 153) and an 80\% held-out test set (n = 608). The calibration set was used for all prompt development and selection, base-rate computation, aggregation weighting and bias-correction fitting; the test set was used only to produce the results reported. Calibration and test sets matched closely in location, dispersion and format composition, except that all five Third-class Cambridge assignments fell, by chance, in the test set.

\subsection{Machine marking}

Three frontier LLMs from independently trained families scored the corpus: GPT-5.4 (OpenAI), Claude Opus 4.6 (Anthropic) and Gemini 3 Flash (Google), each the most capable model from its provider that, at the time of scoring (March 2026), met the project's data-governance requirements (UK GDPR and the project's ethics approval) and sustained stable structured-output access at corpus scale. For Google, the latter excluded larger, more rate-constrained Gemini variants (Supplementary Note S1). All were accessed through their native APIs at temperature 0. Each model scored each assignment in a single, stateless, zero-shot pass -- no exemplar assignments, fine-tuning or iterative refinement -- returning a structured response containing a percentage mark on the institution's permitted scale, a self-reported confidence (uninformative, and excluded from analysis; Supplementary Note S5), and a predicted probability distribution over the permitted marks, used for aggregation.

The zero-shot design matches the adoption route under evaluation -- staff prompting general-purpose models directly, currently the most common form in which these systems are entering marking practice (Attewell, 2025b; Freeman, 2025) -- and tests the evaluative capacity of the model itself rather than that of a scoring system supervised on local exemplars or fine-tuned on local marks (Ke \& Ng, 2019; Latif \& Zhai, 2024; Yancey et al., 2023), whose selection is itself a known source of large performance variance (Lu et al., 2022; Perez et al., 2021; Zhao et al., 2021).

\subsection{Prompt design and selection}

Prompt sensitivity was assessed on the calibration set with a fully crossed 3 × 3 × 3 factorial design over three properties of a marking prompt, each operationalising a documented influence on machine scoring (Jonsson \& Svingby, 2007; Liu et al., 2023; Sadler, 2009; Stureborg et al., 2024; Weigle, 2002). Criterion specificity (factor A) ranged over no rubric (A1), condensed grade boundaries with the institution's criterion names (A2), and the institution's full rubric (A3). Calibration intervention (factor B) was cumulative: none (B1), grade base rates computed from the calibration set (B2), and base rates plus an explicit instruction to counter the tendency of language-model markers to compress marks into a narrow mid-range (B3). Scoring strategy (factor C) contrasted direct scoring (C1), analytic scoring criterion by criterion (C2), and band-first scoring, selecting a degree-class band before a mark within it (C3). Each level was operationalised by one fixed text (full prompt texts in Supplementary Note S2). Because base rates were computed from volunteered submissions, they may differ from the institution's full cohort.

Each model scored the calibration set under all 27 configurations, and factorial analyses used the 146 calibration assignments with complete marks across all 81 model × configuration combinations. For the main evaluation, each model was run under the configuration minimising root-mean-squared error (RMSE) against human marks on the pooled calibration set -- A1-B2-C3 for Claude, A1-B2-C2 for Gemini and A2-B3-C2 for GPT -- applied identically at all three institutions, with only institution-specific content (institution name, rubric text, criterion names, base rates) varying. RMSE was preferred to QWK for its stability at calibration-sample sizes and to avoid optimising the headline agreement metric, and selection was pooled because per-institution optima proved unstable (Supplementary Table S4).

\subsection{Aggregation}

Combining several models is the standard remedy proposed for the unreliability of any one (Verga et al., 2024), but the gain available from any combination rule is bounded by how decorrelated the judges' errors are (Bates \& Granger, 1969; Clemen, 1989; Krogh \& Vedelsby, 1995), and concerns about algorithmic monoculture suggest that frontier models may share their errors (Bommasani et al., 2022); this is tested directly in Section 4.3. We combined the three models' marks under four strategies: an unweighted mean and a median, using the marks alone; an inverse-RMSE weighted mean, scaling each model's contribution by its calibration accuracy at the assignment's institution (Cooke, 1991); and an exploratory meta-prediction variant of the Surprisingly Popular algorithm (Prelec et al., 2017), selecting the mark with greatest mass in the models' averaged predicted distributions. Because averaging suppresses idiosyncratic error, ensemble associations with linguistic features are expected to exceed individual models' on statistical grounds alone, so ensemble and individual-model results are interpreted separately.

\subsection{Evaluation metrics and reliability}

Agreement was assessed on three complementary scales: rank agreement (Spearman's ρ); calibrated categorical agreement, through quadratic weighted kappa (QWK) and normalised QWK (nQWK), which divides QWK by its maximum given the two marginal distributions and so isolates rank-order agreement from distributional mismatch; and absolute error, through mean absolute error (MAE), root mean squared error (RMSE) and mean signed bias (machine minus human). Three coarser statistics reflect the discrete marking schemes: exact agreement (EA, identical marks), adjacent agreement (AA, within three marks) and band agreement (BA, same UK degree classification band: First, 70 and above; upper second, 60--69; lower second, 50--59; Third, 40--49; Fail, below 40). Because the corpus carries single moderated marks, human inter-marker benchmarks are drawn from the published studies cited in Section 2.4. Test--retest reliability was assessed by rescoring a subset of 165 assignments, stratified over institution × degree-class band, five times per model under identical settings (Supplementary Note S4).

\subsection{Linguistic features analysis}

For each assignment we extracted 33 features: 32 linguistic features drawn from prior literature on automated writing assessment (Arnold et al., 2018; Etaat, 2026; Mizumoto \& Eguchi, 2023), spanning lexical diversity and sophistication, readability, syntactic and structural complexity, discourse and cohesion, and writing error (Crossley, 2020), together with word count (definitions and extraction tools in Supplementary Table S8). On the held-out test set (n = 608), we computed Holm-corrected Spearman correlations between the features and the human, individual-model and ensemble marks, pooled and within institution; reduced the standardised features by principal component analysis, retaining six components (79.8\% of feature variance) by the scree-plot elbow criterion; and regressed each mark on the component scores, with institution as a dummy-coded covariate (Cambridge as reference) so that the linguistic coefficients reflect within-institution variation.

\subsection{Statistical analysis}

All analyses were conducted in Python, and all tests were two-sided. Factorial effects were tested by repeated-measures ANOVA on absolute error within model, with partial η² effect sizes; because sphericity cannot be assumed, only effects surviving the conservative lower-bound correction are interpreted. Holm correction was applied within two pre-specified families: nine per-model prompt-property main effects, and fifteen planned paired contrasts for the rubric and calibration interventions, with Wilcoxon signed-rank tests (Wilcoxon, 1945) as distribution-free checks. Inter-model concordance was measured by ICC(2,k) and Krippendorff's α (Krippendorff, 2011), and error dependence was tested through partial correlations between models' signed errors, controlling for the human mark, and the frequency of same-direction errors relative to the independence expectation. Between-institution MAE gaps were decomposed into grade-composition and within-band-accuracy components with a two-fold Kitagawa--Oaxaca decomposition (Kitagawa, 1955; Oaxaca, 1973). Bias corrections (an additive shift, and an affine rescaling matching the human mean and variance) were fitted on the calibration set and transferred unchanged to the test set. Differences between methods (ΔMAE, ΔQWK) were assessed by paired bootstrap over test-set assignments (1,000 resamples, percentile 95\% CIs).

\section{Results}

\subsection{Machine marks agree only moderately with human marks, and aggregation does not reliably improve accuracy}

Each model was evaluated through a model-specific prompt configuration selected on the pooled calibration set (n = 153, 20\% of assignments) and held fixed across institutions, with institution-specific content (institution name, criterion names, grade base rates, and rubric text where the selected configuration included one) resolved within that fixed design (Section 3.3). All accuracy results are computed on the pooled held-out test set (n = 608, 80\%). Model outputs were combined through four aggregation strategies: an unweighted mean, an inverse-RMSE weighted mean, a median, and a meta-prediction variant of the Surprisingly Popular algorithm (Section 3.4).

Because each institution marks against a fixed set of permissible grades, we evaluate agreement using chance-corrected ordinal measures (quadratic weighted kappa, QWK, and its normalised variant, nQWK) and raw accuracy at three levels of stringency (exact, adjacent and band agreement: EA, AA, BA), alongside rank correlation (Spearman's ρ), absolute error (MAE and RMSE, in grade points) and bias (mean signed error). Definitions are given in Section 3.5, with formal definitions in Supplementary Note S1.

\begin{figure}[tbp]
\centering
\includegraphics[width=\linewidth]{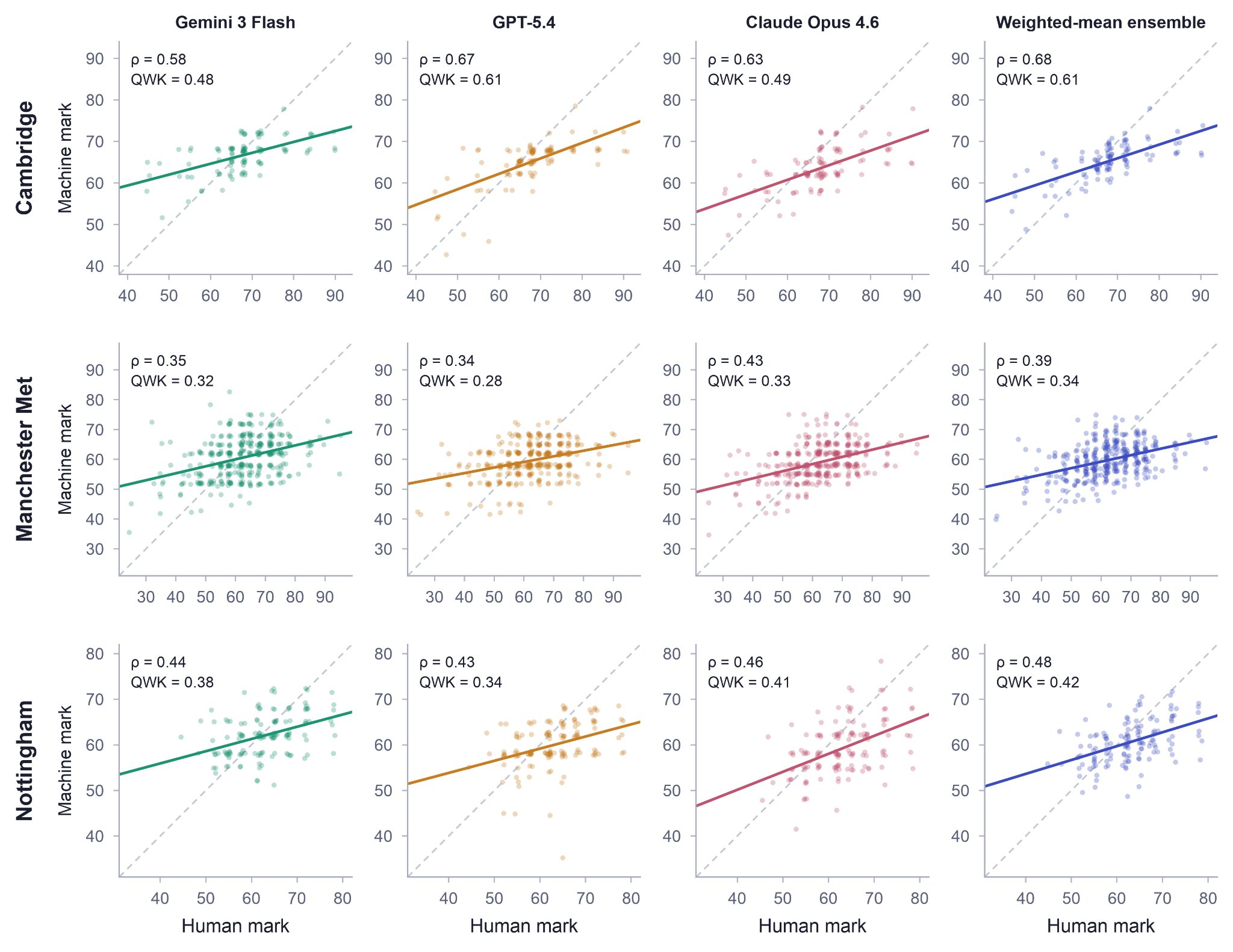}
\caption{AI models track human marking only moderately, and every model compresses the grade scale. Machine marks plotted against human marks for each model and the inverse-RMSE weighted-mean ensemble (columns) at each institution (rows). Each point is one assignment from the held-out test set (Cambridge n = 106; Manchester Met n = 364; Nottingham n = 138), scored in a single zero-shot pass under the model's best-calibrated prompt configuration. Solid lines are least-squares fits and dashed lines mark the identity. Panel insets report Spearman's ρ and QWK. Fitted slopes are shallower than the identity line in every model--institution combination, indicating that machine marks systematically overvalue the weakest assignments and undervalue the strongest.}\label{fig:1}
\end{figure}

Agreement with human marking varied widely across institutions (Fig. 1; Table 1). The strongest method reached a QWK of 0.616 at Cambridge, but only 0.420 at Nottingham and 0.345 at Manchester Met. Within each institution, no single model dominated: GPT was strongest at Cambridge (QWK = 0.609; ρ = 0.671) but weakest at Manchester Met on rank and chance-corrected agreement; Claude led at Manchester Met on both rank agreement and MAE, and at Nottingham on rank agreement; Gemini was the least biased model at every institution and the most accurate at Nottingham on MAE. Judged against human inter-marker correlations documented for comparable material (0.47--0.93), machine--human rank agreement at Cambridge (ρ = 0.67 for the strongest model) sits within the range human markers achieve with one another, whereas at Manchester Met and Nottingham it falls towards, or below, the lower bound of that range. Because the best-performing model differs by institution and by metric, the choice of model cannot be made in advance: it can only be established empirically in the context where deployment is intended.

Aggregation did not reliably improve accuracy. The weighted-mean ensemble was statistically indistinguishable from the best individual model at every institution (ΔQWK = −0.002 at Cambridge, 95\% CI {[}−0.062, 0.074{]}; +0.010 at Manchester Met {[}−0.041, 0.055{]}; +0.011 at Nottingham {[}−0.043, 0.072{]}; paired bootstrap over test-set assignments). The ensemble's value is instead insurance: because the best model varies by institution, averaging delivers near-best performance without a model-selection decision that institutional variation makes unreliable. The two averaging rules (``Mean'' and ``W. Mean'' in Table 1) performed near-identically (QWK differences ≤ 0.009 at every institution) and both outperformed the median. The Surprisingly Popular variant (``SP'') underperformed both at every institution, indicating that the meta-prediction signal added no accuracy beyond what averaging captures. We report the inverse-RMSE weighted mean as the reference ensemble throughout (pooled results for all methods in Supplementary Note S3 and Table S5).

\begin{table}[tbp]
\centering
\small
\setlength{\tabcolsep}{5pt}
\begin{tabular}{@{}lrrrrrrrrr@{}}
\toprule
\textbf{Method} & \textbf{$\rho$} & \textbf{EA (\%)} & \textbf{AA (\%)} & \textbf{BA (\%)} & \textbf{QWK} & \textbf{nQWK} & \textbf{MAE} & \textbf{RMSE} & \textbf{Bias} \\
\midrule
\multicolumn{10}{@{}l}{\textit{Cambridge (n = 106)}} \\
Claude    & .631 & 14.2 & 37.7 & 54.7 & .493 & .672 & 6.3 & 8.2 & $-$4.2 \\
Gemini    & .579 & 31.1 & 49.1 & 62.3 & .484 & .617 & 5.1 & 7.5 & \textbf{$-$1.1} \\
GPT       & .671 & \textbf{32.1} & 49.1 & 58.5 & .609 & \textbf{.742} & \textbf{5.0} & \textbf{7.3} & $-$2.6 \\
Median†   & .666 & 27.4 & 48.1 & 60.4 & .568 & .716 & 5.1 & \textbf{7.3} & $-$2.5 \\
Mean†     & \textbf{.679} & 10.4 & 50.0 & 61.3 & \textbf{.616} & .719 & 5.2 & \textbf{7.3} & $-$2.7 \\
SP†       & .596 & 25.5 & 49.1 & 58.5 & .483 & .651 & 5.4 & 7.7 & $-$2.4 \\
W. Mean†  & .678 & 9.4 & \textbf{50.9} & \textbf{63.2} & .607 & .719 & 5.2 & \textbf{7.3} & $-$2.5 \\
\midrule
\multicolumn{10}{@{}l}{\textit{Manchester Met (n = 364)}} \\
Claude    & \textbf{.433} & \textbf{15.1} & \textbf{28.6} & \textbf{36.8} & .334 & \textbf{.438} & \textbf{8.3} & 10.6 & $-$3.4 \\
Gemini    & .351 & 10.2 & 25.5 & 34.3 & .318 & .363 & 8.6 & 10.9 & \textbf{$-$1.8} \\
GPT       & .336 & 11.5 & 26.4 & 35.4 & .282 & .363 & 8.5 & 10.9 & $-$2.8 \\
Median†   & .371 & 12.4 & 24.7 & 34.9 & .302 & .381 & 8.4 & 10.7 & $-$2.7 \\
Mean†     & .393 & 4.4 & 25.0 & 34.9 & \textbf{.345} & .424 & \textbf{8.3} & \textbf{10.5} & $-$2.7 \\
SP†       & .363 & 11.8 & 25.8 & 36.0 & .306 & .382 & 8.4 & 10.8 & $-$2.9 \\
W. Mean†  & .390 & 4.4 & 25.3 & 34.9 & .344 & .423 & \textbf{8.3} & \textbf{10.5} & $-$2.7 \\
\midrule
\multicolumn{10}{@{}l}{\textit{Nottingham (n = 138)}} \\
Claude    & .456 & 17.4 & 37.0 & 47.8 & .409 & \textbf{.499} & 5.9 & 7.6 & $-$3.7 \\
Gemini    & .436 & \textbf{21.0} & 44.9 & \textbf{56.5} & .378 & .423 & \textbf{5.1} & \textbf{6.7} & \textbf{$-$0.8} \\
GPT       & .430 & 17.4 & 38.4 & 47.8 & .345 & .429 & 5.7 & 7.6 & $-$2.9 \\
Median†   & .462 & 20.3 & 41.3 & 51.4 & .397 & .486 & 5.3 & 6.9 & $-$2.5 \\
Mean†     & \textbf{.481} & 5.1 & 39.9 & 50.7 & .415 & .487 & 5.3 & 6.9 & $-$2.5 \\
SP†       & .439 & 18.1 & \textbf{45.7} & 52.9 & .375 & .450 & 5.3 & 7.0 & $-$2.2 \\
W. Mean†  & \textbf{.481} & 7.2 & 40.6 & 52.9 & \textbf{.420} & .490 & 5.2 & \textbf{6.7} & $-$2.3 \\
\bottomrule
\end{tabular}
\caption{Machine--human scoring agreement is moderate, uneven across institutions, and no single model dominates. Scoring accuracy by method and institution on the held-out test set. Each method was evaluated on the identical set of assignments and against the identical human mark, so all within-block comparisons are paired. Complete cases only (Cambridge n = 106, Manchester Met n = 364, Nottingham n = 138). Individual models (Claude Opus 4.6, Gemini 3 Flash, GPT-5.4, abbreviated by provider) were evaluated under their best-calibrated prompt configuration (see Method). Boldface font indicates the best value per column within each institution (for bias, smallest absolute value). † denotes aggregation methods combining the three models\textquotesingle{} marks; ρ, Spearman correlation; W, inverse-RMSE weighting. EA, exact agreement (identical mark); AA, adjacent agreement (within 3 marks); BA, band agreement (same UK degree classification band, i.e. First, 2:1, 2:2, Third, Fail); QWK, quadratic weighted kappa; nQWK, QWK normalised against the maximum achievable given the two marginal distributions; MAE and RMSE, mean absolute and root mean squared error in grade points; Bias, mean signed error (negative = machine marks below human marks).}\label{tab:1}
\end{table}

\subsection{Marks depend on prompt formulation, and no prompt is optimal across models or institutions}

To measure how far machine marks depend on their elicitation, we crossed three properties of a marking prompt in a 3 × 3 × 3 factorial design (Fig. 2a): criterion specificity (no rubric; condensed grade boundaries; full rubric), calibration intervention (none; the expected grade distribution supplied as base rates; base rates plus an explicit debiasing instruction), and scoring strategy (direct; analytic decomposition by criterion; deliberative band-first). Each model scored the calibration set (n = 153, while the ANOVA used the 146 complete cases) under all 27 configurations; the configuration minimising RMSE on the pooled calibration set was carried forward, unchanged, to the held-out test set (full design and prompt texts in Supplementary Note S2).

Prompt choice alone moved a single model's calibration error by up to a third of a degree classification, but the strength of this sensitivity was a property of the model, not of prompt design itself. Gemini, the most prompt-sensitive model, spanned RMSEs of 8.6 to 11.9 grade points across configurations (3.3 points), Claude spanned 10.5 to 11.7 (1.2 points), GPT only 8.8 to 9.5 (0.7 points) (Supplementary Table S1). Consistently, repeated-measures ANOVA on absolute error (assignments as cases) attributed the largest share of variation to model identity (F(2, 290) = 19.9, p = 8 × 10⁻⁹, ηp² = .121) and to its interactions with the design, which rivalled or exceeded the design main effects: model × criterion specificity (ηp² = .064) and model × scoring strategy (ηp² = .035) both exceeded their corresponding main effects (.053 and .015). Only effects surviving the conservative lower-bound sphericity correction are treated as established; the model × calibration intervention interaction (ηp² = .019) does not (p = .10) and is not interpreted (Supplementary Table S2). Localising these interactions (per-model main effects Holm-corrected as a family of nine, planned contrasts as a family of fifteen, Supplementary Tables S2--S3), each model was most sensitive to a different prompt property (Fig. 2b--d): Gemini to criterion specificity (corrected p = 1 × 10⁻⁴, ηp² = .076; Fig. 2b), Claude to scoring strategy (corrected p = 2 × 10⁻⁴, ηp² = .072; Fig. 2d), and GPT to none -- its largest effect (scoring strategy, ηp² = .023) did not survive correction (corrected p = .16).

Base rates improved accuracy for Claude and Gemini (mean \textbar error\textbar{} −0.46 and −0.48; t(145) = −3.09 and −3.11; both corrected p = .024) but not for GPT (+0.01, corrected p ≈ 1; Fig. 2c), whose marks they instead pushed substantially further downward (mean signed error −0.90, t(145) = −4.93, corrected p = 3 × 10⁻⁵), worsening its existing under-marking. The explicit debiasing instruction improved no model: it worsened Claude's accuracy (+0.21, t(145) = 3.60, corrected p = .005) and pushed two models' marks further downward (signed error: Claude −0.32, corrected p = 6 × 10⁻⁷; GPT −0.22, corrected p = .035; Gemini +0.19, corrected p ≈ 1) -- the wrong direction, as all three models already under-marked on average.

Two structural properties of this sensitivity matter more than its magnitude. First, the optimal elicitation belonged to the model, not the task: no configuration was optimal across models, and the optimal prompt differed for every model. Second, no optimum transferred across institutions (Supplementary Table S4): selecting configurations separately at each institution yielded no shared optimum within any model -- Gemini's pooled optimum (no rubric) was inverted at Cambridge, where its best configuration of 27 used the full rubric -- and configuration rankings were inconsistently correlated between institutions (Spearman ρ: −.61 to +.53, median ≈ 0), indicating that with the current number of complete calibration assignments per institution (27 for Cambridge, 87 for Manchester Met and 32 for Nottingham), local optima may reflect sampling noise, the very instability that motivated pooled selection (Section 3.3).

Rubric interpretation was itself model-dependent: the full rubric substantially worsened Gemini's accuracy (mean \textbar error\textbar{} +1.26 points, t(145) = 3.98, corrected p = .001; marginal RMSE 10.9 vs 9.4; Fig. 2b), left Claude and GPT indifferent (both \textbar Δ\textbar{} ≤ 0.06, corrected p ≈ 1), helped no model -- and none of the three selected configurations included it: Claude\textquotesingle s and Gemini\textquotesingle s used no rubric at all, and GPT\textquotesingle s used only the condensed grade boundaries.

\begin{figure}[tbp]
\centering
\includegraphics[width=\linewidth]{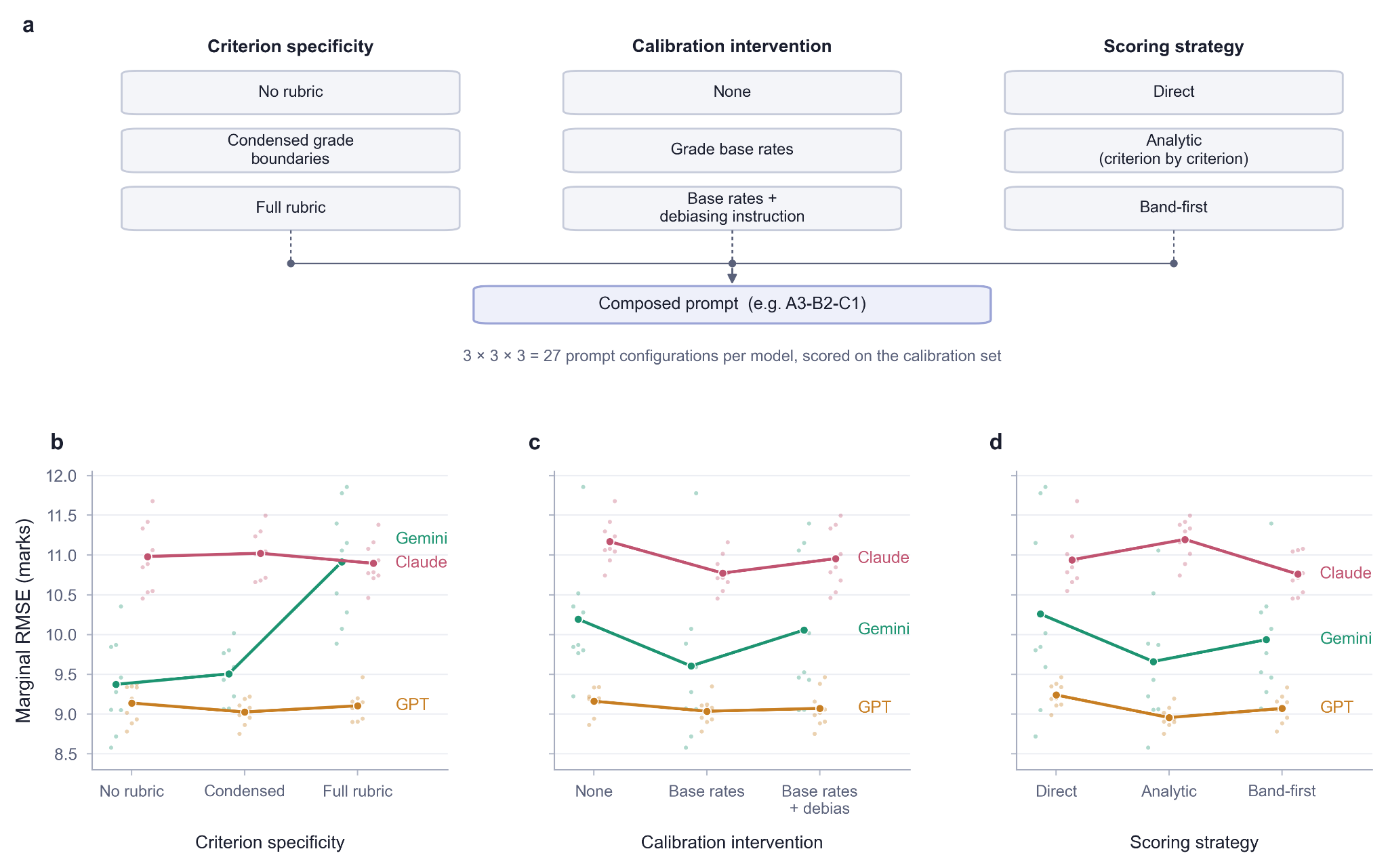}
\caption{Effects of prompt design on marking accuracy are model-specific. (a) The fully crossed 3 × 3 × 3 prompt design: three levels each of criterion specificity, calibration intervention, and scoring strategy, composing 27 prompt configurations per model, scored on the calibration set. (b--d) Marginal root mean squared error (RMSE, in grade points, lower is better) on the calibration set (n = 153, 146 complete cases, assignments across all 81 model--configuration combinations), pooled across institutions, for each prompt property: (b) criterion specificity; (c) calibration intervention; (d) scoring strategy. Each point is a model's RMSE marginalised over the other two factors (nine configurations per point). The three models diverge in which factor they are sensitive to: Gemini's error rises sharply with the full rubric (b) and falls under analytic scoring (d), Claude benefits from the band-first strategy (d), and GPT is close to indifferent across all levels. Base rates help Claude and Gemini, but the added debiasing instruction reverses the gain (c).}\label{fig:2}
\end{figure}

\subsection{Machine marking errors are shared across models and only partially correctable}

Underlying the moderate and uneven agreement documented above is a single error structure that every model reproduced at every institution: compression of the grade scale towards a common attractor (pooled ensemble mean = 60.9; per-model means 59.8--62.0, Fig. 3a). Machine marks exhibited substantially less dispersion than human marks, with standard deviations well below those of human marks in every model × institution cell (0.46--0.83 of the human s.d., Fig. 3b). Regression of machine on human marks yields slopes far below unity (Claude b = 0.30, 95\% CI {[}0.25, 0.34{]}; Gemini b = 0.27 {[}0.23, 0.32{]}; GPT b = 0.25 {[}0.21, 0.29{]}; ensemble b = 0.27 {[}0.23, 0.31{]}; all p \textless{} 0.001 against a unit slope), reflecting the joint effect of this reduced dispersion and imperfect machine--human correlation. In practical terms, a 10-point difference between two assignments translated into a difference of only 2.5--3 points in their predicted marks.

Compression made model error highly predictable from the human mark alone: signed error correlated with the human mark at r = −0.79 to −0.83 (Fig. 3a), so where an assignment sat on the grade scale explained 62--69\% of error variance. The consequences of compression were clearest at the degree-class boundaries. Across institutions, assignments awarded 60--64 by human markers were scored with little bias (−1.7 points; MAE = 3.8). Bias then increased in both directions: marks in the 50s were over-marked by +2.9 points, those awarded 65--69 were under-marked by −3.9 points, and the tails of the distribution were missed by more than 11 points on average (+11.6 below 50; −11.9 for First-class assignments, with MAE exceeding 11 points at both extremes). Compression therefore distorted grades most severely exactly where summative decisions carry the greatest consequences. On top of this compression sat a consistent negative offset, with mean signed bias ranging from −0.8 to −4.2 grade points -- smallest for Gemini (−0.8 at Nottingham), largest for Claude (−4.2 at Cambridge).

Establishing that this error structure is genuinely shared, rather than an artefact of scoring noise, requires two steps. First, run-to-run variability is excluded: rescoring a fixed set of 165 assignments five times showed near-deterministic within-model behaviour (test--retest ICC(2,1) = 0.968--0.999; degree band stable on 86.7--99.4\% of assignments; Supplementary Note S4), so the marks each model produces, and the errors they contain, are systematic and not stochastic. Second, despite belonging to three different model families, the three models produced highly concordant marks (pooled ICC(2,k) = 0.91; Krippendorff's α = 0.77; Supplementary Table S6). Concordance of marks partly reflects shared signal, so the better test is whether the models' errors covary, and they do. Pairwise correlations between the models' signed errors remained substantial after partialling out the human mark (partial r = 0.74--0.78; raw r = 0.90--0.93, inflated by the human mark shared between both error terms). All three models erred in the same direction on 63.4\% of assignments (43.3\% all below the human mark, 20.1\% all above) -- 3.4 times the 18.8\% that would be expected were their errors independent, given each model's marginal error tendencies. Consistent with this, inter-model agreement had little relationship to accuracy: it was highest at Manchester Met (ICC = 0.92), where performance against human markers was weakest, and lowest at Nottingham (ICC = 0.85). High consensus among models did not signal correctness. Because the benefit of aggregation is bounded by how decorrelated its members' errors are, this error structure explains why ensembling failed to remove both biases.

\begin{figure}[tbp]
\centering
\includegraphics[width=\linewidth]{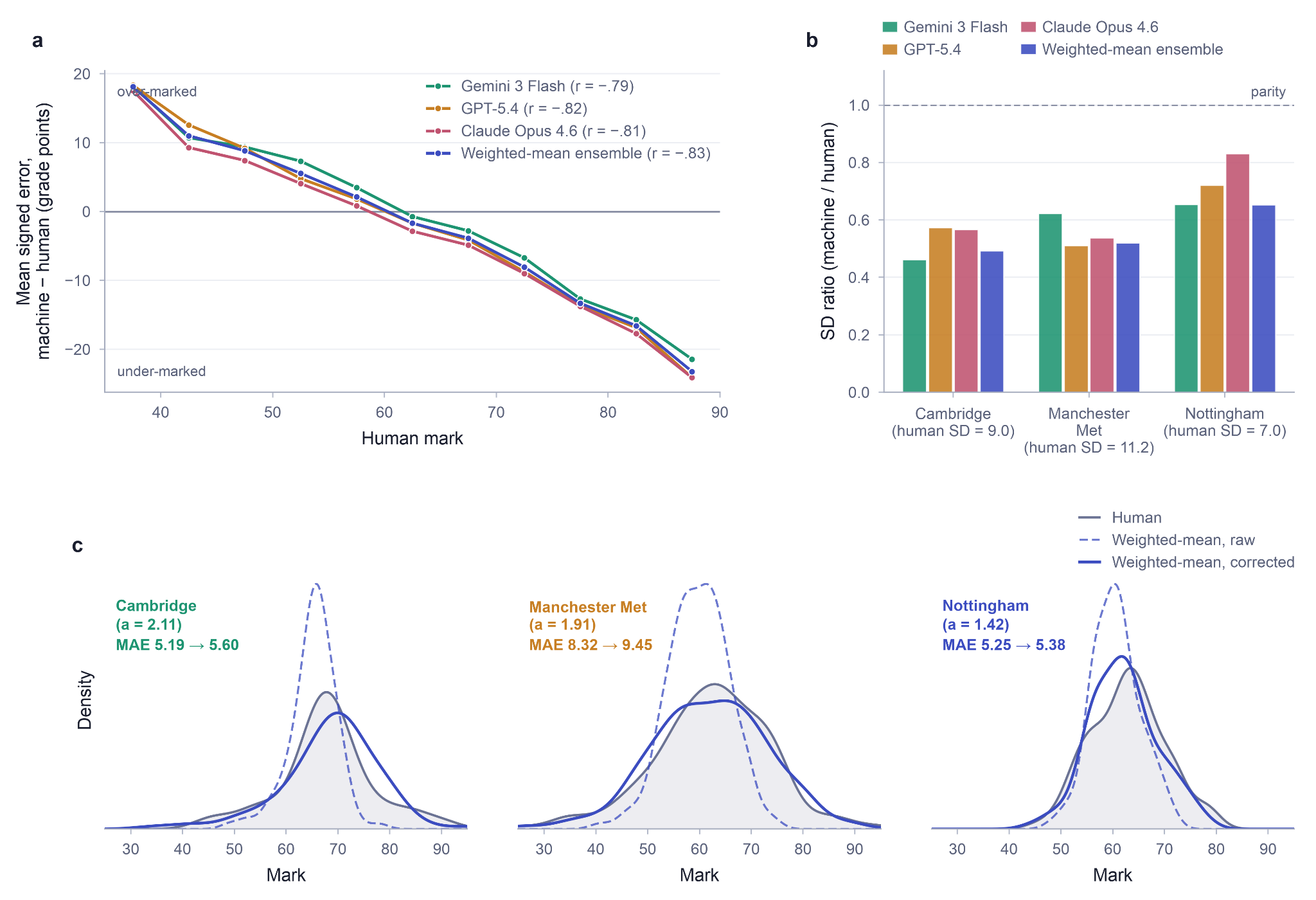}
\caption{Shared compression towards a common central tendency resists linear calibration. (a) Mean signed error (machine − human, grade points) as a function of the human mark. Errors cross zero near the lower boundary of the 2:1 band and grow monotonically towards both tails for every model (correlations between signed error and human mark: −0.79 to −0.83), such that assignment position on the grade scale alone explains 62--69\% of error variance. (b) Ratio of machine to human mark standard deviations by institution. All cells fall below parity (dashed line), spanning 0.46 (Gemini, Cambridge) to 0.83 (Claude, Nottingham). (c) Density of human marks against raw and affine-corrected weighted-mean predictions at each institution. Matching the calibration set's mean and variance requires stretch factors of 2.11 (Cambridge), 1.91 (Manchester Met) and 1.42 (Nottingham); the correction restores the spread of the marks but amplifies assignment-specific residuals alongside the signal, trading absolute accuracy for chance-corrected ordinal agreement while leaving rank agreement unchanged.}\label{fig:3}
\end{figure}

As the dominant error structure was systematic, we next asked whether it could be removed through post hoc calibration. Two shared biases were identified: the consistent offset and the more consequential compression of the grade scale. We therefore evaluated simple linear corrections for each. The offset proved substantially easier to correct than the compression. A single additive adjustment per model, estimated on the calibration set and transferred unchanged to the test set, improved accuracy meaningfully for the most biased models. For Claude, a +3.9-point adjustment increased pooled QWK from 0.393 to 0.459 (bootstrap 95\% CI for ΔQWK = {[}0.032, 0.086{]}) while reducing MAE from 7.39 to 6.71 (ΔMAE = {[}−0.94, −0.43{]}); GPT and the weighted mean showed significant MAE reductions of approximately 0.37 points; and Gemini, already the least biased model, showed no reliable improvement.

Compression, by contrast, could not be eliminated by linear calibration. Restoring the human grade distribution required rescaling predictions away from the attractor with an affine transformation matching the calibration set's mean and variance, stretching predictions by factors of 1.4--2.1 across institutions (Fig. 3c). This improved QWK by restoring the spread of the marks, but it amplified assignment-specific prediction error, because every residual was stretched alongside the signal. For Claude, the affine correction increased QWK beyond the shift-only adjustment (0.393 to 0.491) while accuracy deteriorated (MAE 7.39 to 7.58). Relative to the additive correction alone, the additional rescaling yielded ΔQWK = +0.032 (95\% CI {[}0.010, 0.057{]}) at a cost of ΔMAE = +0.87 ({[}0.60, 1.15{]}). The trade-off was steeper for the weighted mean, whose larger gain in QWK (+0.043 {[}0.017, 0.073{]}) came with a larger 1.15-point increase in MAE ({[}0.80, 1.50{]}). Linear calibration can therefore remove the shared offset, but the further transformation that restores the spread of the marks increases chance-corrected ordinal agreement at the price of average error: beyond the offset correction, the two metrics pull in opposite directions.

\subsection{Agreement between human and machine marking does not generalise across institutions}

Because every model compresses marks in the same way, a similar error profile appears at every institution (Fig. 1): pooled across institutions, signed errors cross zero near the lower boundary of the UK 2:1 band, become increasingly positive below it, and increasingly negative above it (Fig. 3a). Were this shared attractor the whole explanation here, machine--human agreement at each institution would be determined simply by where its marks lie: distributions concentrated near the compression point should agree well, distributions with mass in the tails should not.

Cross-institutional variation was indeed substantial, and its ordering partly fits that prediction. Cambridge, whose high-centred grade distribution (M = 67.7, SD = 9.0) concentrates over half its marks within the upper-second band (Fig. 4a), showed by far the highest agreement (best QWK = .616): 53\% of its test assignments lie in the 60s, where the ensemble errs by only 2.4 points on average. Agreement was weakest at Manchester Met (best QWK = .345), where the distribution was wider (SD = 11.2) and 64\% of assignments lie outside the 60s, including 26\% in the First-class range, where the ensemble under-marks by 13.8 points on average. Yet distributional position and width alone cannot explain the ordering: Nottingham's distribution is the narrowest of the three (SD = 7.0) and centred lower than Cambridge's (M = 62.9), yet its agreement remains intermediate (best QWK = .420).

If these performance differences arose solely because institutions assign different mixtures of grades, then reweighting one institution's grade distribution to match another's should largely eliminate the gap. Yet it does not. A Kitagawa decomposition of the 3.1-point MAE difference between Manchester Met and Cambridge (Fig. 4b) attributes only 0.4 points (12\%; 15\% under finer 5-point binning) to grade composition, with the remaining 2.7 points (88\%) arising from differences in within-band accuracy. Manchester Met assignments are marked less accurately across almost the entire grade scale, with MAE in the 60s alone more than twice Cambridge's (5.4 vs 2.4), the sparsely populated 50s band (n = 10 at Cambridge) being the only exception. Nottingham illustrates the same point from the opposite direction (Fig. 4c): despite an overall MAE matching Cambridge's, its substantially lower QWK (.420 vs .607 for the same ensemble) shows that differences in grade distribution still degrade rank agreement even when mean accuracy is unchanged.

\begin{figure}[tbp]
\centering
\includegraphics[width=\linewidth]{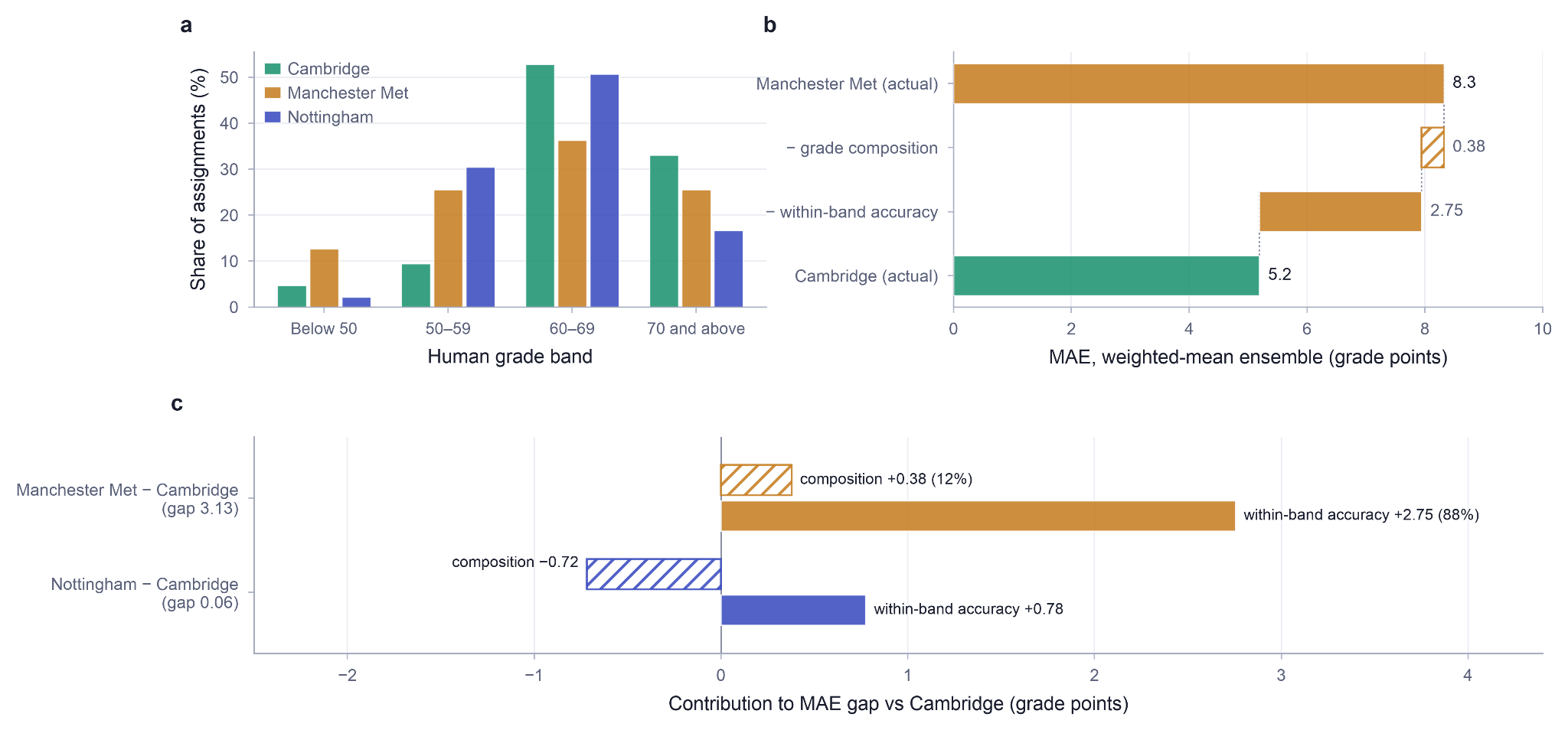}
\caption{Machine--human agreement fails to generalise across institutions mainly through within-band accuracy, not grade composition. (a) Distribution of human marks by degree band at each institution. (b) Kitagawa decomposition of the weighted-mean ensemble's mean absolute error gap between Manchester Met (8.3 grade points) and Cambridge (5.2): equalising cohort composition removes only 0.38 points of the difference; the remaining 2.75 points arise from lower within-band agreement between machine and human marks. (c) Contributions of grade composition and within-band accuracy to each institution's MAE gap relative to Cambridge. Nottingham's overall MAE is similar to Cambridge's only because a favourable composition (−0.72 points) offsets within-band accuracy that is 0.78 points worse. Composition explains approximately one-eighth of the Manchester Met--Cambridge gap. The remainder reflects institution-specific structure, in the assignments, the writing, or the human benchmark itself, that is invisible to single-institution evaluation.}\label{fig:4}
\end{figure}

The failure of performance to generalise across institutions therefore stems from two different sources. The first is compositional: institutions differ in where their students' marks lie relative to the models' shared attractor, so the same error profile translates into different agreement statistics. The second lies in the model × institution interaction: the same models, under identical configurations, marked different institutions' assignments with systematically different accuracy within the same grade bands. Because the models were held fixed, this variation may plausibly originate in what differs between institutions -- assignments, marking conventions, student writing (though not assessment genre, which did not predict absolute scoring error within institutions; Supplementary Note S6) -- but it manifests only through the models' differential sensitivity to those differences, an interaction our design detects but cannot further decompose.

\subsection{The evidential basis diverges: machine marks are influenced by features that human marks discount}

To understand the divergence documented above, we examined how far human and machine marks were conditioned on the measurable surface properties of the assignments. We extracted 33 features per assignment: 32 linguistic features spanning five domains established in the automated scoring literature -- lexical diversity, readability, syntactic and structural complexity, discourse and cohesion, and writing error -- together with word count (Supplementary Note S7; Supplementary Table S8).

At the bivariate level, human and machine marks already diverge in what they track. Holm-corrected Spearman correlations between marks and features showed the same directional pattern for human and machine marks, but the two differed markedly in breadth and strength. Human marks correlated reliably with five features, concentrated in lexical sophistication, readability and writing error, whereas the weighted-ensemble machine mark showed significant associations with 20 of the 33 (Supplementary Fig. S1). Re-estimating the correlations within each institution (Fisher z-transformed to permit comparison of effect sizes) showed that the asymmetry is itself context-dependent. Ensemble correlations exceeded human correlations in magnitude for 29 of 33 features at Manchester Met and 27 of 33 at Cambridge, but for only 14 of 33 at Nottingham (Fig. 5). The broader conditioning of machine marks on the linguistic surface is therefore robust in the pooled analysis but not uniform across cohorts.

\begin{figure}[tbp]
\centering
\includegraphics[width=\linewidth]{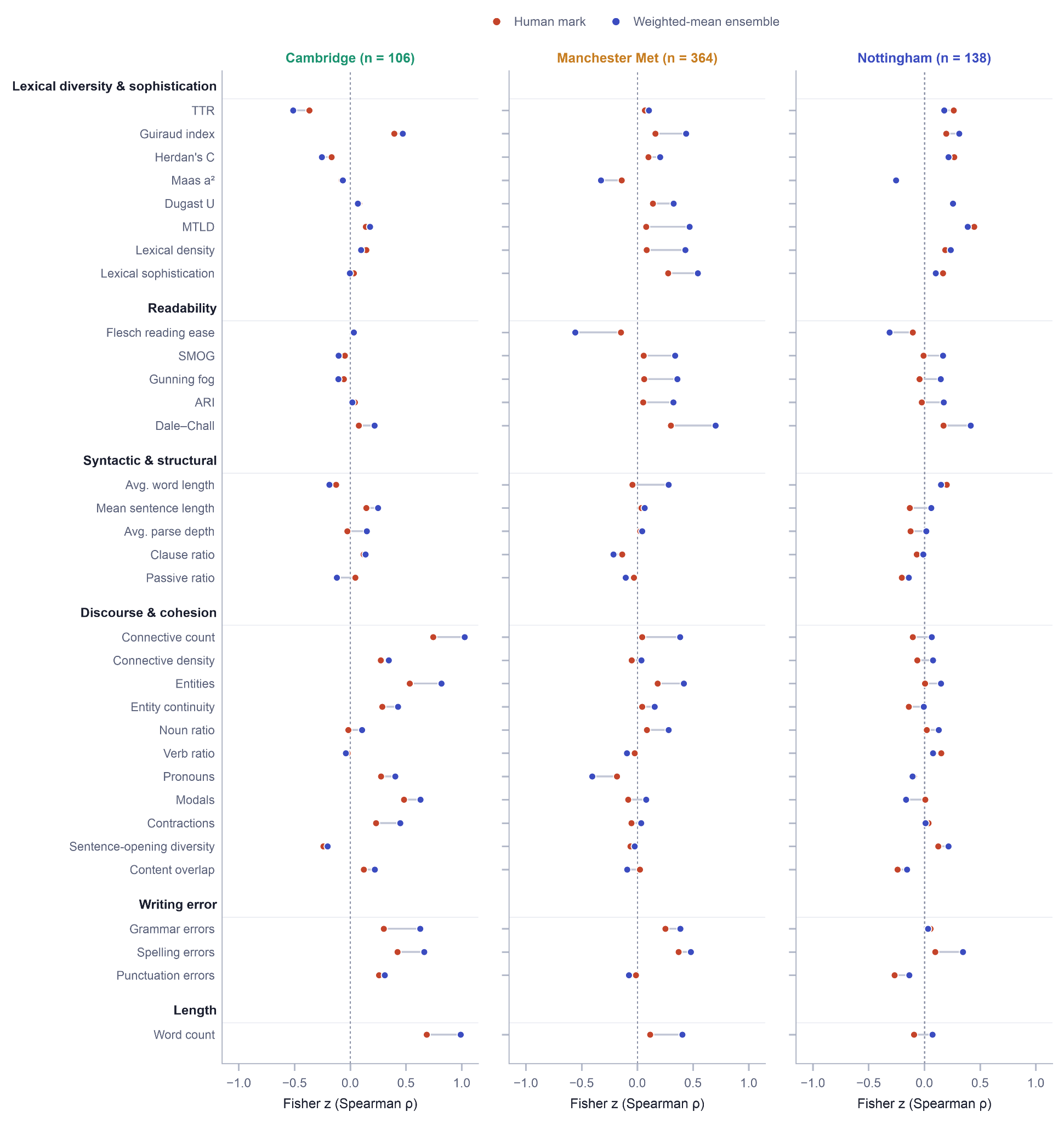}
\caption{Machine marks are more strongly conditioned on surface linguistic features than human marks, and feature sensitivity varies by institution. Fisher z-transformed Spearman correlations between marks and each of the 33 extracted features (rows, grouped by linguistic domain) at each institution (columns), for human marks (red) and the weighted-mean ensemble (blue); connecting lines show the human--ensemble gap for each feature. Sample sizes are the held-out test set (Cambridge n = 106; Manchester Met n = 364; Nottingham n = 138). Ensemble correlations exceed human correlations in magnitude for 29 of 33 features at Manchester Met and 27 of 33 at Cambridge, but for only 14 of 33 at Nottingham, and the profile of feature--mark associations differs substantially across institutions, indicating that the influence of surface features on marking is context-dependent. Feature definitions are given in Supplementary Table S8.}\label{fig:5}
\end{figure}

Because the features intercorrelate substantially -- several measures capture overlapping constructs rather than distinct properties of writing -- individual feature--mark associations are an incomplete guide to the underlying structure. We therefore reduced the 33 standardised features by principal component analysis, retaining six components that jointly explained 79.8\% of feature variance (24.1\%, 19.2\%, 16.2\%, 10.7\%, 6.8\% and 2.8\% respectively) and captured interpretable dimensions of vocabulary use, sentence complexity and readability, length and information content, cohesion, and stylistic register (Supplementary Fig. S2; component retention in Section 3.6).

Linear models regressing each mark on the six components, with institution as a covariate, quantified the divergence (Table 2). Assignment length and information density (PC3) was the strongest predictor of both human marks (b = 0.98, SE = 0.19, p \textless{} .001) and ensemble machine marks (b = 0.84, SE = 0.09, p \textless{} .001), and lexical richness (PC1) predicted both (human b = 0.58, SE = 0.14; ensemble b = 0.73, SE = 0.06; both p \textless{} .001). The critical dissociation concerned sentence complexity and readability (PC2): it reliably predicted the ensemble mark (b = 0.34, SE = 0.08, p \textless{} .001) but had no relationship to the human mark (b = 0.01, SE = 0.17, p = .954). No other component reliably predicted either mark (all p ≥ .051).

The models differed most, however, in overall dependence on the linguistic surface: the six components and institutional covariates together explained 38.5\% of variance in ensemble machine marks against 11.1\% in human marks (F(8, 599) = 46.78 and 9.36 respectively, both p \textless{} .001). Part of this gap is expected on statistical grounds alone, since averaging across models suppresses idiosyncratic error and thereby sharpens any shared conditioning on measurable features (Section 3.4). The per-model regressions in Supplementary Table S9 show that the asymmetry holds, attenuated, for each model individually (R² = 0.320--0.365, with PC2 a significant predictor for every model). Machine marks are therefore not merely noisier versions of human judgements: they are substantially more predictable from properties of the writing, and along at least one dimension -- the sentence-complexity component, PC2 -- they are conditioned on a property that carried no detectable weight in human marking.

\begin{table}[tbp]
\centering
\footnotesize
\setlength{\tabcolsep}{4pt}
\begin{tabular}{@{}llrrrrrrr@{}}
\toprule
 & & & \multicolumn{3}{c}{\textbf{Human mark}} & \multicolumn{3}{c}{\textbf{Ensemble mark}} \\
\cmidrule(lr){4-6}\cmidrule(l){7-9}
\multicolumn{2}{@{}l}{\textbf{Component}} & \textbf{Var. (\%)} & \textbf{\textit{b}} & \textbf{SE} & \textbf{\textit{P}} & \textbf{\textit{b}} & \textbf{SE} & \textbf{\textit{P}} \\
\midrule
PC1 & Lexical richness \& diversity & 24.1 & \textbf{0.58} & \textbf{0.14} & \textbf{$<$0.001} & \textbf{0.73} & \textbf{0.06} & \textbf{$<$0.001} \\
PC2 & Sentence complexity \& readability & 19.2 & 0.01 & 0.17 & 0.954 & \textbf{0.34} & \textbf{0.08} & \textbf{$<$0.001} \\
PC3 & Assignment length \& information density & 16.2 & \textbf{0.98} & \textbf{0.19} & \textbf{$<$0.001} & \textbf{0.84} & \textbf{0.09} & \textbf{$<$0.001} \\
PC4 & Lexical density vs cohesion & 10.7 & $-$0.11 & 0.22 & 0.601 & 0.01 & 0.10 & 0.887 \\
PC5 & Conversational vs formal style & 6.8 & $-$0.52 & 0.26 & 0.051 & 0.14 & 0.12 & 0.255 \\
PC6 & Informal style & 2.8 & 0.14 & 0.41 & 0.729 & $-$0.11 & 0.19 & 0.567 \\
\midrule
\multicolumn{3}{@{}l}{\textit{Model fit}} & \multicolumn{3}{r}{$R^2 = 0.111$} & \multicolumn{3}{r}{$R^2 = 0.385$} \\
 & & & \multicolumn{3}{r}{$F(8, 599) = 9.36$} & \multicolumn{3}{r}{$F(8, 599) = 46.78$} \\
 & & & \multicolumn{3}{r}{$P < 0.001$} & \multicolumn{3}{r}{$P < 0.001$} \\
\bottomrule
\end{tabular}
\caption{Machine marks are conditioned on a linguistic dimension undetected in human marks, and on the linguistic surface overall more strongly. Linear regressions of human and ensemble machine marks on the six PCA-derived linguistic components, held-out test set (n = 608). Var. (\%) gives the percentage of variance in the feature set explained by each component (cumulative 79.8\%), not variance in marks. Unstandardised coefficients (b) with standard errors and two-sided p values; institution was included as a dummy-coded covariate (Cambridge as reference; covariate coefficients in Supplementary Table S9). Boldface indicates coefficients significant at p \textless{} .05. Sentence complexity and readability (PC2) predicts the ensemble mark but not the human mark.}\label{tab:2}
\end{table}

\section{Discussion}

Across five dimensions of evaluation, our results suggest that although frontier LLMs can reach moderate agreement with human marking, that agreement is a local property of a particular model, prompt and cohort, not evidence of a transferable evaluative capacity. Agreement depended partly on where a cohort's marks lay relative to the region of the grade scale where the models' shared compression produces little error, but differed between institutions even within the same grade bands; it depended on elicitation choices whose effects were themselves model-specific and institution-specific; it could not be improved by aggregation, because the models' errors were correlated; and it reflected separate evidential bases, where machine marks were more strongly conditioned on linguistic features than human marks.

The compression of the grade scale deserves particular attention, because it is not a defect that can straightforwardly be removed. It appeared in every model from three independent providers and survived prompting interventions designed to counteract it, extending to multiple models and institutions the central tendency previously reported for single models in smaller studies (Lundgren, 2024; Wetzler et al., 2025). This consistent behaviour of machine marking -- accurate near the centre of the distribution, increasingly wrong towards both tails -- undermines its readiness for deployment in real-world assessments, because degree outcomes often turn on precisely the tail cases that the models miss most. A machine marker can be right on average and still be wrong wherever it matters.

The divergent evidential basis may help explain why. The construct-validity literature warned, long before LLMs, that automated systems could match human marks while rewarding assignment length and lexical sophistication rather than the constructs assessed (Perelman, 2014; Powers et al., 2002; Williamson et al., 1999). Indeed, our results indicate that the linguistic components, together with institution, explained roughly three times as much variance in each model's marks as in human marks, rising to three and a half times for the ensemble. The divergence is largely one of degree rather than of kind: the direction of feature--score relationships was consistent across human and machine evaluation, with both sensitive to lexical richness and to assignment length and information density. The exception is sentence complexity and readability, which carried detectable weight only in machine marking. Machine marking thus amplifies signals human markers already use while adding one that carried no detectable weight in human marks -- a distinction with remedial consequences, since amplified signals can be reweighted whereas a novel criterion must first be detected.

Meanwhile, most of the variance in human marks remained unexplained by measurable linguistic properties, consistent with human assessors drawing on qualitative, contextual and discipline-specific judgement that standard computational measures of writing do not capture. This raises fairness concerns: surface linguistic features covary with language background and educational history (Cheng \& Furnham, 2019; Crawford et al., 1988), so any systematic reward for them redistributes marks along lines unrelated to attainment. This also creates a perverse incentive to write for the model rather than the discipline (O'Neil, 2016), and as writing tools improve (Noy \& Zhang, 2023), a marker conditioned on style will discriminate least well among the work it would increasingly judge.

High consensus among models did not signal correctness: inter-model agreement was highest at the institution where accuracy against human markers was lowest. Panels of LLMs are therefore not independent witnesses whose consensus approximates truth, but correlated instruments whose consensus can entrench a common bias behind the appearance of corroboration. The practical insight for current institutional experimentation is that verifying one model's mark with a second model, or a panel, may reinforce, not cancel, the shared compression.

\subsection{Implications for institutions}

For institutions, the implications are procedural as much as evaluative. Because performance did not transfer across three universities within a single discipline and educational system, and because only around one-eighth of the largest cross-institutional accuracy gap was attributable to grade composition, validation conducted elsewhere, however rigorous, licenses nothing locally. Any contemplated deployment requires local evidence, built on the institution's own assessments and marking practices and repeated whenever the underlying model changes.

\subsection{Limitations}

Our design supports strong claims about context-dependence but bounded ones about scope. The corpus comprises psychology assignments from three UK universities. Other disciplines, assessment genres and grading systems may yield different profiles, although the mechanisms we identify -- compression towards a learned prior and sensitivity to surface features -- are unlikely to be discipline-specific. Each assignment carries a single moderated institutional mark rather than multiple independent re-marks, so human--human agreement is estimated from external evidence rather than measured within-corpus, and attenuation of machine--human agreement by criterion noise cannot be fully separated from true divergence. We evaluated zero-shot marking under a finite (though systematically crossed) prompt space. Fine-tuned or exemplar-calibrated models may behave differently and merit the same scrutiny. Our findings attach to specific model versions accessed at a fixed time, and vendor updates can alter behaviour silently, which is also an argument for continuous local monitoring rather than one-off validation. And our feature analysis is correlational: it establishes that machine marks are conditioned on surface properties human marks discount, not the internal mechanism by which this occurs. These limitations qualify the breadth of our conclusions, not their direction: every one of them counsels more caution about deployment, not less.

\section{Conclusion}

Frontier LLMs can approximate human marking in some settings, but the agreement we observed was a joint product of model, prompt and institution rather than evidence of an evaluative capacity that transfers across contexts. Every model compressed the grade scale towards a common central tendency, the models shared rather than cancelled one another's errors, and machine marks leaned more heavily than human marks on surface features of the writing. Single-context validation therefore overstates readiness for deployment: institutions considering machine marking need evidence built on their own assessments, renewed whenever the underlying model changes.

\section*{Acknowledgements}

We would like to thank our Advisory Board (C Cheung, G Chikwa, K Crockett, H Spain, C Kuepper-Tetzel, S Watson, C Zapata, Y Garg, N Franyo, R Meijer, and I Bobrovsky) as well as I Fay, SN Mattey, L Razzaq, and H Tal for their support throughout this project.

\section*{Funding}

AM, DT, GC, MAT, RT, and YB disclose support for the research from ai@cam and the Accelerate Programme for Scientific Discovery, made possible by a donation from Schmidt Sciences.

\section*{Contributions}

GC conducted all analyses except the linguistic analyses and wrote the associated code. MAT carried out the linguistic analyses, wrote the corresponding code and drafted its description. GC, MAT and GTA carried out data collection and data curation, and GC and MAT were responsible for visualisation. AM, DT and RT supervised the main methodology, data collection and associated analyses. AM, DT, GC, MAT, RT and YB conceived the study, developed the methodology and provided key intellectual input. GTA administered the project. AM, DT, MAT, RT and YB acquired funding. GC wrote the first draft of the manuscript with contributions from AM and MAT. All authors provided edits and feedback, and read and commented on the final manuscript.

\section*{Declaration of competing interest}

The authors declare no competing interests.

\section*{Ethics statement}

The procedures in this research project were approved by the University of Cambridge Higher Education Studies Research Ethics Committee (CHESREC), reference 2025.LT.84. Written informed consent was obtained from all students whose assignments were included in the corpus. Students were informed that participation was voluntary, that consent could be withdrawn without academic consequence, and that assignment texts would be pseudonymised before analysis and before any submission to third-party model providers. Assignment texts and marks were stored on institutional infrastructure and linked only by pseudonymous identifiers. No identifiable participant information is reproduced in this manuscript.

\section*{Data and code availability}

To enable scrutiny and reuse, we provide the complete de-identified analytical dataset -- human marks, machine marks under every prompt configuration and model, extracted linguistic features and institutional identifiers -- at \url{https://zenodo.org/records/21676733}. These materials permit full reproduction of all reported results. Note that the student assignments analysed in this study cannot be made publicly available. The assignments are the intellectual property and personal data of their student authors, who consented to their use by the research team for this project only. Public release would be incompatible with the terms of consent granted to the project by the University of Cambridge Higher Education Studies Research Ethics Committee (CHESREC). The associated examination and coursework questions likewise cannot be released, as they form part of live assessment banks at the three participating universities and their disclosure would compromise the security and integrity of future examinations. The full codebase and replication scripts are available at \url{https://github.com/giuliocorsi/machine-human-marking-agreement}.

\section*{Declaration of generative AI and AI-assisted technologies in the manuscript preparation process}

During the preparation of this work the authors used Claude Code (Anthropic) to assist with writing analysis and figure-generation code, and Claude (Anthropic) to review the manuscript text for clarity, spelling and grammar. After using these tools, the authors reviewed, tested and edited the code and content as needed and take full responsibility for the content of the published article.

\clearpage
\setcounter{figure}{0}
\setcounter{table}{0}
\setcounter{equation}{0}
\renewcommand{\thefigure}{S\arabic{figure}}
\renewcommand{\thetable}{S\arabic{table}}
\captionsetup[figure]{name=Supplementary Fig.}
\captionsetup[table]{name=Supplementary Table}

\begin{center}
{\LARGE Supplementary Material}\\[0.75em]
{\large \papertitle}
\end{center}
\bigskip

\section*{Supplementary Note S1: Full methods}

This note gives the complete methods, of which Section 3 of the main text is a condensed version. The ethics statement is given in Section 3.1 of the main text.

\subsection*{Corpus and human marks}

The full corpus comprised 934 marked undergraduate psychology assignments from three UK universities: the University of Cambridge (n = 248), Manchester Metropolitan University (n = 461), and the University of Nottingham (n = 225). All were summative coursework or examination online submissions marked by academic staff at each institution as part of routine assessment processes for the degree. Marking used each institution's established assessment criteria and the standard UK percentage scale (0--100), although only some numbers on this scale were `allowed' within each institution's guidelines. We excluded 172 short-form assessment items a priori (115 Cambridge, 4 Manchester Met, 53 Nottingham), as we considered these to represent a different assessment genre from assignment-length work. Classification as short form was based on the assessment type rather than the length of the student response. This distinction was important because some assignment-length assessments elicited relatively short responses despite requiring extended-answer formats.

Of the remaining 762 assignments, one Manchester Met assignment was excluded because one model returned no mark for it owing to an API error. The two marks that were returned for this assignment were extreme low outliers, approximately 50 points below the human mark, consistent with a processing failure rather than an evaluative judgement. This left the complete-case corpus of 761 (Cambridge n = 133, Manchester Met n = 456, Nottingham n = 172), in which every assignment carries a human mark and a valid mark from all three models. All sample sizes refer to this subset.

The corpus spans 50 modules and 87 distinct assignments: 45 examination assignment questions at Cambridge and 42 coursework assignments at Manchester Met and Nottingham. Assessment formats comprise examination assignments (Cambridge, n = 133), coursework assignments (Manchester Met n = 239, Nottingham n = 106), quantitative and qualitative research reports (Manchester Met n = 122, Nottingham n = 20), case-study analyses (Manchester Met n = 54), applied scenario responses (Nottingham n = 41), reflective assignments (Manchester Met n = 35), and research proposals and portfolios (Manchester Met n = 6, Nottingham n = 5). Assignment length averaged 1,325 words at Cambridge (range 451--2,368), 1,730 at Manchester Met (753--3,082), and 1,467 at Nottingham (135--4,068).

Institutional marking schemes are discrete: marks are permitted only at fixed points (ticks) on the percentage scale (17 permitted marks at Cambridge and Nottingham, 32 at Manchester Met). Within the UK upper-second band (60--69), for example, all three schemes permit only 62, 65, and 68, and the schemes differ mainly in the tails, where Cambridge and Nottingham become sparse while Manchester Met remains uniform. Assignments were assessed against each institution's published criteria, which all included range and command of sources; argument, critical analysis and interpretation; and style, structure and presentation.

Mark distributions of contributed assignments differed markedly across institutions (complete-case corpus: Cambridge M = 68.0, s.d. = 8.6, range 45--90; Manchester Met M = 62.5, s.d. = 11.2, range 25--95; Nottingham M = 62.6, s.d. = 6.8, range 45--78); test-set values are reported in Section 4.4 of the main text. All assignments were de-identified before processing. Models were not provided with any student-identifying information.

\subsection*{Calibration and test splits}

The 762 assignment-length submissions were partitioned once, deterministically and pseudo-randomly within institution (seeded pseudo-random assignment), into a 20\% calibration set (153 assignments: 27 Cambridge, 92 Manchester Met, 34 Nottingham) and an 80\% test set (n = 609). The subsequently excluded assignment fell, by chance, in the test set, so no quantity estimated on the calibration set was affected and the held-out test set comprises 608 complete assignments. Pseudo-random within-institution assignment preserves that distribution in expectation, and did so in practice: the calibration sets matched their test sets closely in location, dispersion, and format composition (means within 1.2 marks, s.d. within 2 marks; e.g.~research reports constitute 25\% of Manchester Met's calibration set and 27\% of its test set), and contained every degree-class band present in the corpus at Manchester Met (First through Fail) and Nottingham (First through Third; no Fail marks occur in Nottingham). The exception is Cambridge, where the five Third-class assignments in the corpus all fell to the test set by chance, so calibration there rests on marks between 52 and 84 against a test-set range of 45--90. Consequently, tail behaviour at Cambridge is uninformed by calibration.

The calibration set was used to estimate institution-level quantities: grade base rates for the calibration interventions and calibration RMSE for the aggregation weights. It was also used for prompt development, prompt selection, aggregation weighting, base-rate computation, and bias-correction fitting. The test set was used only to produce the results reported. No test assignment informed any prompt, weight, or correction at any stage.

\subsection*{Machine marking}

Three frontier LLMs scored the corpus: GPT-5.4 (OpenAI), Claude Opus 4.6 (Anthropic) and Gemini 3 Flash (Google), each amongst the most capable models of its provider meeting the criteria below at the time of scoring (March 2026). Model selection applied three criteria. First, one model from each of three independently trained families, to reduce the risk of correlated errors. Second, data governance: because the corpus comprises student work, processing was restricted to providers offering a series of guarantees that submissions would not be retained or used for training, under terms compliant with UK GDPR and the project's ethics approval. This excluded providers unable to offer such terms at the time, irrespective of capability. Third, engineering: within each provider, the selected model's API had to sustain stable structured-output access at corpus scale throughout the scoring window, which excluded some larger, more rate-constrained variants of the same families. All models were accessed through their native provider APIs at temperature 0, so that prompt construction, response parsing and grade extraction were identical across models.

Each model scored each assignment in a single zero-shot pass with no persistent memory -- no exemplar assignments, fine-tuning or iterative refinement -- returning a structured response containing three fields: a percentage mark over the institution's permitted scale; a self-reported confidence, elicited for completeness but excluded from all analysis after calibration-set checks showed it to be uninformative (Supplementary Note S5); and a predicted probability distribution over the institution's permitted ticks, used by the Surprisingly Popular aggregation.

The zero-shot design is a deliberate choice, for four reasons. First, it matches the adoption route under evaluation: assessment staff prompting general-purpose models directly -- the most accessible and, on current survey evidence, the most common form in which these systems are entering marking practice (Attewell, 2025b; Freeman, 2025) -- operate zero-shot by construction, so a zero-shot evaluation tests the proposal in the form in which it is actually arriving. Second, the alternatives would have answered a different question. Supplying exemplar scripts with their marks demonstrably improves agreement in some settings (Yancey et al., 2023), but it does so by converting a general-purpose model into a locally supervised scoring system -- a return, in practice, to classical automated essay scoring with an LLM backbone (Ke \& Ng, 2019), whose performance attests to the calibration data rather than to any evaluative capacity of the model itself. It is the latter that our design interrogates. Third, the two supervised alternatives fail on data at authentic scale, in different ways. Fine-tuning requires volumes of marked scripts per institution that authentic corpora such as this one rarely provide.

Exemplar-based prompting requires far fewer scripts, but introduces a further discrete selection problem -- which exemplars, how many, spanning which bands, per institution -- that is known to produce large performance variance, with even the ordering of a fixed exemplar set moving models between near-optimal and near-chance accuracy (Lu et al., 2022; Zhao et al., 2021), and that cannot be reliably optimised on small validation sets without overestimating performance (Perez et al., 2021). Fourth, a single stateless pass per assignment preserves construct purity: every mark reflects one prompt and one script, with no conversational state through which configurations, or prior scripts, could contaminate scores. Within the zero-shot regime, elicitation was nonetheless explored systematically through the factorial design described below, and the calibration interventions supply the distributional information that constitutes the informational core of what exemplar sets provide.

\subsection*{Factorial prompt design}

Prompt sensitivity was assessed with a fully crossed 3 × 3 × 3 factorial design over three properties of a marking prompt, tested on the calibration set only. Each property operationalises a documented influence on machine scoring: criterion specificity addresses the effects of rubric presence and granularity on both human and machine markers (Jonsson \& Svingby, 2007; Stureborg et al., 2024); calibration intervention targets the compression of scores into a narrow mid-range documented in LLM evaluators (Stureborg et al., 2024); and scoring strategy contrasts holistic with decomposed judgement, mirroring the analytic and band-based scales of human writing assessment (Sadler, 2009; Weigle, 2002), and structured evaluation in LLM scoring (Liu et al., 2023).

Criterion specificity (factor A) ranged over no rubric (A1), condensed grade boundaries -- each band's mark range with a one-line quality descriptor, plus the institution's criterion names (A2), and the institution's full rubric with per-band descriptors (A3). Calibration intervention (factor B) was cumulative: none (B1), grade base rates (B2), and base rates plus an explicit debiasing instruction, warning that language-model markers tend to compress marks into a narrow mid-range and directing the model to use the full scale where the evidence warrants (B3). Scoring strategy (factor C) contrasted direct scoring (C1), analytic scoring -- evaluating the submission criterion by criterion against the institution's named criteria before synthesising a single mark (C2), and band-first scoring -- selecting a degree-class band before a mark within it (C3). Both structured strategies carried an identical instruction to conduct their reasoning internally and exclude it from the output, so that scoring strategy was not confounded with output verbosity.

Base rates (B2, B3) comprised marginal band frequencies only, with no anchoring exemplars, computed solely from each institution's calibration set -- that is, from the distribution of volunteered submissions rather than the institution's full cohort. Because participation was voluntary, these distributions may differ from cohort-level mark distributions, and differentially so across institutions.

The design enforced four construct-purity rules: a single task instruction appeared identically in all 27 configurations; prompt-block order was fixed (task, rubric, distribution, debiasing, strategy, output format). The output-format instruction was defined once and never modified, and each factor level was operationalised by exactly one fixed text, so that conclusions attach to these operationalisations rather than to the constructs at large. Every assignment × configuration × model combination was scored in a separate, stateless API call containing only that prompt and that assignment. No conversation state was shared between calls, so no configuration could influence scoring under any other.

Each model scored the full calibration set under all 27 configurations. Factorial analyses used the 146 calibration assignments (27 Cambridge, 87 Manchester Met, 32 Nottingham) with complete marks across all 81 model × configuration combinations, giving a balanced repeated-measures design. The initial size of the calibration set was 153, and seven calibration assignments lacked a mark in at least one cell of the factorial grid and were therefore excluded.

\subsection*{Prompt selection for the main evaluation}

For the main evaluation, each model scored the entire corpus under its calibration-selected prompt configuration -- the configuration minimising root-mean-squared error (RMSE) against human marks on the pooled calibration set: A1-B2-C3 for Claude (no rubric, base rates, band-first), A1-B2-C2 for Gemini (no rubric, base rates, analytic) and A2-B3-C2 for GPT (condensed boundaries, base rates plus debiasing, analytic). Selection used RMSE rather than QWK for three reasons. First, RMSE is a smooth function of continuous error and therefore stable at calibration-sample sizes, whereas QWK (computed from a contingency table over grade categories and normalised by chance agreement) is volatile in small samples with clustered marks, exactly the conditions of the calibration set. Second, RMSE's quadratic penalty weights most heavily the large errors at the scale's extremes, where misclassification is most consequential. Third, and perhaps most importantly, selecting on a criterion distinct from the headline agreement metric guards against overfitting the reported statistic: configurations were chosen for error minimisation, and their test-set QWK is therefore not directly optimised by the selection procedure. The selected configuration was applied identically at all three institutions. Only institution-specific content (rubric text, criterion names, base rates) varied within it. Selection was performed on the pooled calibration set rather than per institution because per-institution optima proved unstable at the available calibration sample sizes.

\subsection*{Aggregation}

Aggregating the three models' marks serves two purposes in the design. Practically, a panel of machine markers is the strongest form of the deployment proposal: whatever procurement route an institution takes -- including single-vendor adoption through existing contracts -- the standard remedy proposed for the unreliability of any individual model is to combine several (Verga et al., 2024), so evaluating LLM marking against a multi-model panel tests the proposal in its most defensible configuration. Theoretically, that proposal inherits the logic of judgement aggregation: where judges err independently, combining them cancels idiosyncratic error and retains shared signal, and the accuracy gain available from any combination rule is bounded by how decorrelated the judges' errors are -- a result long established for forecast combination and ensemble learning (Bates \& Granger, 1969; Clemen, 1989; Krogh \& Vedelsby, 1995) and underlying performance-weighted aggregation in structured expert judgement (Cooke, 1991). Whether frontier models from distinct providers satisfy this independence condition is itself one of the study's questions. Concerns about algorithmic monoculture suggest they may not (Bommasani et al., 2022), and it is tested directly in Section 4.3 through the pairwise correlation of the models' signed errors and the frequency of same-direction errors relative to the independence expectation. The aggregation strategies below therefore function both as candidate scoring methods and as instruments for that test. Treating the three models as a single three-member panel, we combined their marks under four aggregation strategies ordered by the information each exploits: the marks alone (unweighted mean; median), demonstrated accuracy on held-out data (inverse-RMSE weighted mean), and the models' beliefs about one another's judgements (the Surprisingly Popular algorithm; Prelec et al., 2017).

Exploiting the least information (the marks alone), the unweighted mean is the canonical crowd estimate, treating all models as equally informative. The median replaces the mean with the middle judgement, offering robustness to a single outlying marker. With \emph{K} = 3, it discards the highest and lowest score entirely. The weighted mean adopts the performance-weighting logic of structured expert elicitation: rather than counting all judges equally, each model's contribution is scaled by its demonstrated calibration accuracy at the assignment's institution. Weight uncertainty was assessed by bootstrap over calibration assignments (1,000 resamples). Estimating weights per institution is compatible with pooling prompt selection: selection is a 27-way discrete choice in which small RMSE differences flip the winner, whereas the weights are a smooth function of calibration RMSE, so calibration-sample noise perturbs them only marginally.

The Surprisingly Popular approach (Prelec et al., 2017) extends crowd aggregation with the judges' beliefs about the judgements of others, in the canonical form recovering answers that are more common than the crowd predicts. Because the canonical vote-versus-prediction statistic presumes a large panel, with \emph{K} = 3 we use a meta-prediction variant operating on the models' predicted tick distributions: the three distributions are averaged, renormalised, and the tick with greatest aggregated mass is selected. Because this variant departs from the canonical statistic, Surprisingly Popular results are interpreted as exploratory.

All aggregated marks are evaluated in Section 4.1 and Supplementary Table S5. Because averaging suppresses idiosyncratic error, ensemble associations with linguistic features are expected to exceed individual models' on statistical grounds alone. Ensemble sensitivity results are therefore reported alongside, but interpreted separately from, individual-model results. The extent to which aggregation can help at all is an empirical question about error structure -- a crowd of markers sharing the same biases gains little from averaging -- and is tested directly in the shared-error analysis in Section 4.3.

\subsection*{Evaluation metrics}

Agreement between machine and human marks is reported on three complementary scales. Rank agreement is tested via Spearman's ρ. Calibrated categorical agreement is tested through quadratic weighted kappa (QWK), treating marks as ordinal categories over the \emph{K} distinct mark values observed in the sample under evaluation, sorted in ascending order. QWK is computed as

\[
\mathrm{QWK} = 1 - \frac{\sum_{i,j} w_{ij}\, O_{ij}}{\sum_{i,j} w_{ij}\, E_{ij}},\qquad w_{ij} = \frac{(i - j)^{2}}{(K - 1)^{2}},
\]

where \emph{i} and \emph{j} index positions in that sorted list, \emph{O} is the observed joint distribution of machine and human marks, \emph{E} is the distribution expected by chance given the two marginals (their outer product), and the squared-distance weights penalise disagreements in proportion to their ordinal severity, so that QWK = 1 indicates perfect agreement and QWK = 0 agreement no better than chance. Normalised QWK (nQWK) divides QWK by the maximum attainable given the two marginal mark distributions, nQWK = QWK/QWK\textsubscript{max}, where QWK\textsubscript{max} is the value obtained under the joint distribution that maximises agreement subject to the observed marginals -- constructed by sorting both marginals and pairing them rank for rank -- thereby isolating rank-order agreement from distributional mismatch: a model that ranks assignments perfectly but compresses their spread attains nQWK = 1 while QWK remains below it.

Absolute error is computed through mean absolute error (MAE) and root mean squared error, \(\mathrm{RMSE} = \sqrt{\tfrac{1}{n}\sum_{i=1}^{n}(\widehat{m}_{i} - m_{i})^{2}}\), where \({\widehat{m}}_{i}\) and \(m_{i}\) are the machine and human marks for assignment \emph{i}, both in marks, together with mean signed bias (machine minus human), negative values indicating machine marks below human marks. Three coarser statistics reflect the discrete marking schemes against which each institution assesses: exact agreement (EA), the proportion of predictions identical to the human mark; adjacent agreement (AA), the proportion within three marks of the human mark; and band agreement (BA), the proportion within the same UK degree classification band (First, 70 and above; upper second, 60--69; lower second, 50--59; Third, 40--49; Fail, below 40). Human inter-marker benchmarks against which machine agreement is situated are drawn from published studies of degree-level marking cited in Section 2.4, as the corpus carries single moderated marks rather than independent double-marks.

\subsection*{Reliability}

Test--retest reliability was assessed on a fixed subset of 165 assignments (46 Cambridge, 70 Manchester Met, 49 Nottingham), stratified over institution × degree-class band so that reliability is assessed across the full mark range rather than the corpus's centre of the distribution. Whereas the calibration--test partition had to preserve each institution's natural grade distribution -- and assignment was therefore delegated to a single seeded pseudo-random draw -- the reliability subset carries no representativeness requirement: its purpose is coverage, so its composition was fixed directly by design, with 11--16 assignments per stratum and the sparsest strata (Third-class assignments at Cambridge and Nottingham) taken in full. The realised composition is given in Supplementary Note S4. Each model rescored the subset five times under identical settings. Within-model reliability is summarised by the standard deviation of repeated marks per assignment and intraclass correlation. Between-model concordance is summarised separately, on the held-out test set, by ICC(2,k) and Krippendorff's α on the pooled marks.

\subsection*{Linguistic features analysis}

For each assignment we extracted 33 features: 32 linguistic features drawn from prior literature on automated writing assessment (Arnold et al., 2018; Etaat, 2026; Mizumoto \& Eguchi, 2023), together with word count. The linguistic features span five domains -- lexical diversity and sophistication, readability, syntactic and structural complexity, discourse and cohesion, and writing error -- reflecting the established finding that writing quality cannot be adequately represented by any single linguistic measure (Crossley, 2020). Extraction used spaCy and NLTK preprocessing in combination with domain-specific tools: \emph{lexicalrichness} for diversity indices, \emph{wordfreq} for lexical sophistication, \emph{textstat} for readability metrics, and \emph{symspellpy} for grammar and spelling detection. Full feature definitions and extraction methods are given in Supplementary Table S8.

Three analyses were conducted, all on the held-out test set (n = 608). First, pairwise Spearman rank-order correlations were computed between the five mark variables (the human mark; the three individual model marks; the weighted-mean ensemble) and the 33 features, with the Holm correction applied across all 703 unique pairwise comparisons among the 38 variables. Correlations were estimated in two stages: first on the pooled test set, to establish overall feature--mark associations, and then within each institution, to examine whether those associations are institution-dependent; for the institutional comparison, coefficients were Fisher z-transformed (z = arctanh(ρ)) to stabilise variance and permit comparison of effect sizes across institutions.

Second, the 33 features were standardised (z-scored) and entered into a principal component analysis. Six components were retained, jointly explaining 79.8\% of feature variance. Retention was determined by visual inspection of the scree plot using the elbow criterion. Component loadings and interpretations are given in Supplementary Fig. S2.

Third, component scores were entered as predictors in two linear regression models, one with the human mark and one with the ensemble machine mark as outcome. Institution was included as a dummy-coded covariate (Cambridge as reference) so that the linguistic coefficients reflect within-institution variation rather than between-institution differences in mark distributions; the institutional coefficients represent adjusted mean differences relative to Cambridge and are reported in Supplementary Table S9. Unstandardised coefficients (b) are reported with standard errors and two-sided p values. Equivalent models were fitted for each individual model's marks (Supplementary Table S9).

\subsection*{Statistical analysis}

All analyses were conducted in Python. The analysis code is available as described in the Data and code availability statement of the main text. All tests were two-sided, with p values reported.

Factorial results were analysed by repeated-measures ANOVA on absolute error over the 3 × 3 × 3 design within model (146 complete-case assignments; 7 assignments had at least one missing cell), with partial η² effect sizes. Because sphericity cannot be assumed, we report only effects that survive the lower-bound correction, the most conservative adjustment available. Effects that did not survive are reported in Supplementary Table S2 and not interpreted.

Multiple-comparison correction used the Holm procedure within two pre-specified families: nine per-model prompt-property main effects (three models × three properties), and fifteen planned contrasts -- for each model, full vs no rubric on absolute error (A3 vs A1), and base rates vs none and debias vs base rates on both absolute and signed error (B2 vs B1, B3 vs B2) -- evaluated by paired t-tests on per-assignment differences, with Wilcoxon signed-rank tests (Wilcoxon, 1945) as distribution-free checks. All fifteen contrasts are reported in Supplementary Table S3. Signed-error contrasts were planned for the calibration interventions because their hypothesised effect is directional.

Marking reliability and inter-model dependence were quantified as follows. Test--retest stability used ICC(2,1) over repeated runs, while inter-model concordance used ICC(2,k) and Krippendorff's α (Krippendorff, 2011). Because concordant marks partly reflect shared signal, error dependence was tested directly: pairwise Pearson correlations between models' signed errors on the test set, partial correlations controlling for the human mark, and the proportion of assignments on which all three models erred in the same direction, compared against the expectation under independent errors computed from each model's marginal error tendencies.

Cross-institution generalisation was analysed with a symmetric two-fold Kitagawa--Oaxaca decomposition (Kitagawa, 1955; Oaxaca, 1973) of between-institution MAE gaps into grade-composition and within-band-accuracy components, computed over the four degree-class bands (below 50, 50--59, 60--69, 70 and above) with midpoint weights, under which the two components sum exactly to the gap; 5-point binning is reported as a robustness check.

Bias correction was fitted on the calibration set and transferred unchanged to the test set: a per-model additive shift equal to the calibration-set mean signed error, and an affine transformation matching the calibration set's human mean and variance (multiplying centred predictions by s = s.d.(human)/s.d.(model)). Uncertainty for comparisons between methods and between corrected and uncorrected marks (ΔMAE, ΔQWK) was quantified by paired bootstrap over test-set assignments (1,000 resamples, percentile 95\% CIs), resampling assignments jointly so that each resample preserves the pairing of methods on the same assignments. Associations between textual features and marks or errors used Spearman correlations with Holm correction across all 703 feature pairs, with institution differences assessed via Fisher z-transformation.

\section*{Supplementary Note S2: Prompt construction, factorial analysis, and configuration selection}

\subsection*{Prompt construction and component texts}

Prompt configurations varied three properties in a full 3 × 3 × 3 factorial: criterion specificity (A: no rubric, condensed rubric, full rubric), calibration intervention (B: none, base rates, base rates with an explicit debiasing instruction), and scoring strategy (C: direct, analytic, band-first). Each block was defined by exactly one fixed text, so the 27 configurations differ only in which blocks are present, never in their wording or order. Factor levels map to blocks as follows: A1 includes no rubric block, A2 the condensed grade boundaries, and A3 the institution's full rubric; B1 includes no calibration block, B2 the grade distribution, and B3 the grade distribution followed by the debiasing instruction; C1 includes no strategy block, C2 the analytic instruction, and C3 the band-first instruction (implemented as the `deliberative' component in the released prompt library).

Only three slots were localised per institution: the condensed and full rubric texts, drawn from each institution's published marking criteria, and the grade distribution, computed from its calibration-set band frequencies (main text, Section 3.3). Within the shared components, placeholders resolved to institutional content: \{\{INSTITUTION\}\} and \{\{SUBJECT\}\} to the institution's name and discipline, \{\{CRITERIA\_LIST\}\} to its named assessment criteria, and \{\{GRADE\_BANDS\}\} to its degree-classification labels. The complete prompt library, including all three institutions' condensed and full rubric texts, is provided in machine-readable form in the code repository. The shared component texts were, verbatim:

\textbf{Preamble.} ``You are an experienced \{\{INSTITUTION\}\} examiner marking undergraduate \{\{SUBJECT\}\} submissions.''

\textbf{Task.} ``Mark the submission below on the institution's 0-100 scale.''

\textbf{Condensed grade boundaries (A2).} Each band's mark range with a one-line quality descriptor, followed by the institution's criterion names. At Cambridge, for example: ``GRADE BOUNDARIES: 70-100: First (excellent range and command of material, original critical analysis); 60-69: Upper Second (good knowledge, organised, clearly argued); 50-59: Lower Second (reasonable competence, may have gaps); 40-49: Third (basic knowledge only, serious deficiencies); 0-39: Fail (irrelevant, ignorant, or superficial). Assessed across 7 criteria: Range and command of material, Argument and analysis, Understanding of the subject, Engagement with the question, Originality and critical treatment of evidence, Use of sources, Planning and completeness.'' The full per-band rubrics (A3) reproduce each institution's complete published marking criteria and are given in the repository.

\textbf{Grade distribution (B2, B3).} ``GRADE DISTRIBUTION: In a reference sample of previously marked submissions from this course, approximately {[}band percentages from the institution's calibration set{]}.'' For example, at Nottingham: ``approximately 9\% received Firsts (70+), 50\% received 2:1s (60-69), 38\% received 2:2s (50-59), and 3\% received Thirds (40-49); no submissions in the sample received Fails (below 40).''

\textbf{Debiasing (B3).} ``IMPORTANT: Large language models used as markers tend to exhibit central tendency bias - compressing scores into a narrow mid-range and underusing the extremes of the scale. Be aware of this tendency in your own judgement. Use the full range of the marking scale where the evidence warrants it.''

\textbf{Analytic scoring (C2).} ``Evaluate the submission separately on each of the institution's assessment criteria: \{\{CRITERIA\_LIST\}\}. Form a judgement for each criterion, then synthesise these judgements into a single overall mark. Carry out this evaluation internally - do not include it in your output.''

\textbf{Band-first scoring (C3).} ``Before assigning a mark, deliberate: (1) identify which grade band (\{\{GRADE\_BANDS\}\}) best characterises this submission's overall quality, (2) weigh the strongest evidence for and against that classification, (3) determine the specific mark within that band. Carry out this evaluation internally - do not include it in your output.''

\subsection*{Factorial analysis, planned contrasts, and configuration selection}

Each model marked the calibration split under all 27 configurations; RMSE against the human mark served as the selection criterion, with selection performed per model on the pooled calibration split and the chosen configuration held fixed across institutions (main text, Section 3.3). Supplementary Table S1 reports the full grid. Calibration RMSE spanned 10.45--11.68 for Claude, 8.58--11.86 for Gemini, and 8.76--9.47 for GPT.

Because the sphericity assumption is unlikely to hold across 27 conditions, inference in Supplementary Table S2 relies on the conservative lower-bound correction, and only effects surviving it are treated as established. Under it, scoring strategy and calibration intervention, together with their interaction, survive for Claude; criterion specificity and calibration intervention, together with a criterion × strategy interaction, survive for Gemini; and no term survives for GPT. In the combined analysis, the model term is the largest single source of variance (ηp² = .121), with significant model × criterion and model × strategy interactions.

The planned contrasts in Supplementary Table S3 give the surviving effects their direction. Base rates reduced absolute error for Claude and Gemini (Δ = −0.46 and −0.48 marks; dz ≈ −0.26) but not for GPT, for which the only reliable consequences were signed -- a downward shift of 0.90 marks with no accuracy gain. The debiasing instruction improved no model, and for Claude it reversed part of the base-rate gain (+0.21 on absolute error) and shifted marks downward (−0.32 signed). The full rubric conferred no advantage over no rubric for Claude or GPT, and increased Gemini's absolute error by 1.26 marks.

Supplementary Table S4 reports the counterfactual of per-institution selection. Local optima coincide with the deployed pooled optimum in two of nine model × institution combinations; the pooled optimum's local rank ranges from 1st to 18th of 27, and Spearman correlations between institutions' full cell orderings are weak or negative (−.607 to .526). With calibration subsets of 27--87 assignments, differences among adjacent cells are small relative to sampling noise, and per-institution selection would risk fitting that noise.

\begin{table}[tbp]
\centering
\caption{Calibration-split RMSE for all 27 prompt configurations, by model. Root-mean-square error against the human mark on the calibration split for each cell of the 3 × 3 × 3 prompt factorial (A: criterion specificity, B: calibration intervention, C: scoring strategy). Bold values mark the pooled optimum for each model (Claude A1-B2-C3; Gemini A1-B2-C2; GPT A2-B3-C2), deployed unchanged across all three institutions.}\label{tab:S1}
\includegraphics[width=\linewidth]{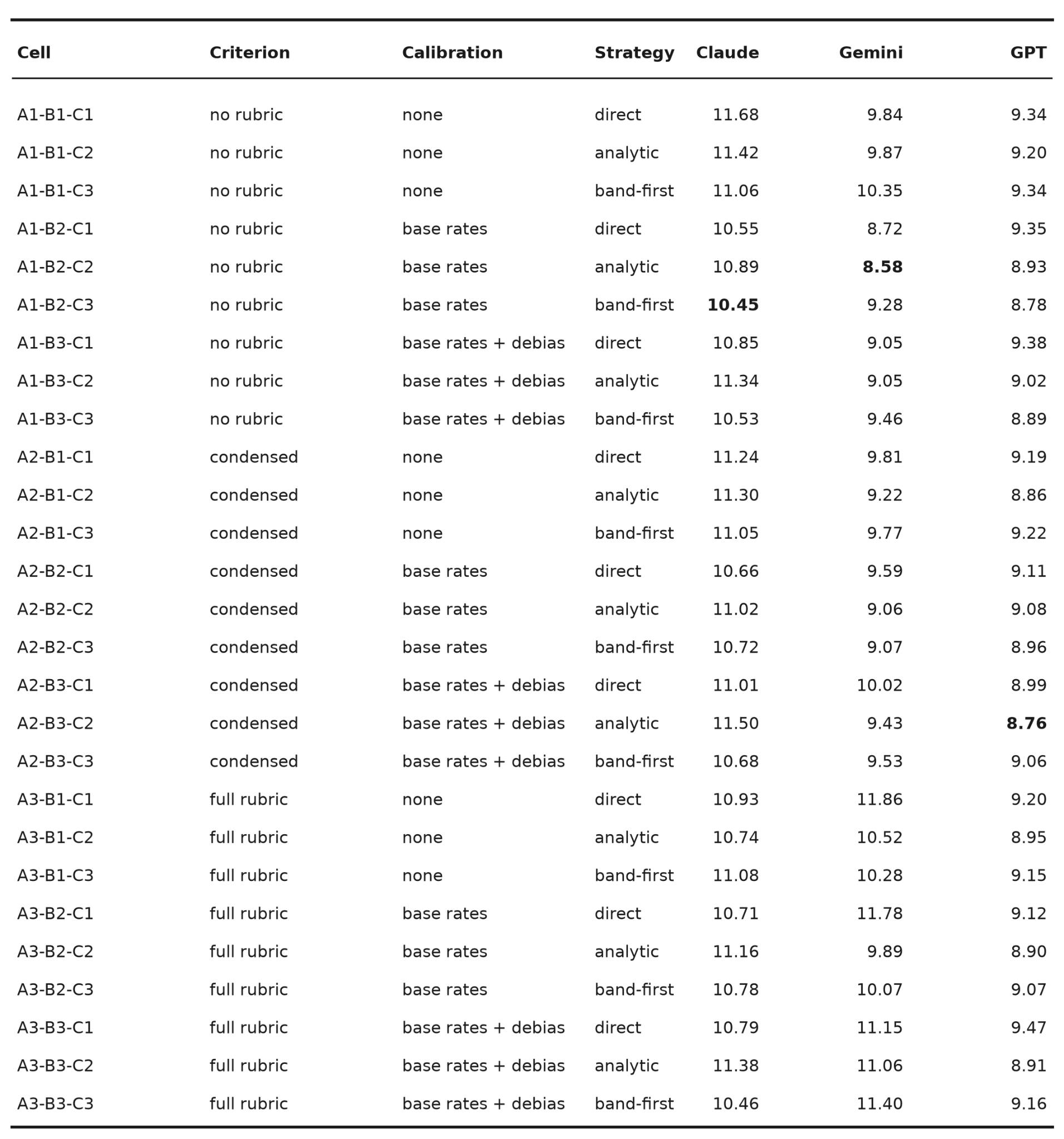}
\end{table}

\begin{table}[tbp]
\centering
\caption{Repeated-measures ANOVA on per-assignment absolute error across the prompt factorial. Within-model analyses treat criterion specificity (A), calibration intervention (B), and scoring strategy (C) as within-assignment factors; the combined analysis adds model as an additional within-assignment factor. p (Holm), Holm-adjusted p values for the pre-specified family of nine per-model prompt-property main effects (three models × three properties). Interactions and the combined analysis were not part of this family and are marked --. p (LB) applies the lower-bound sphericity correction. An effect is recorded as surviving only if the corrected p \textless{} .05. ηp², partial eta squared.}\label{tab:S2}
\includegraphics[width=0.75\linewidth]{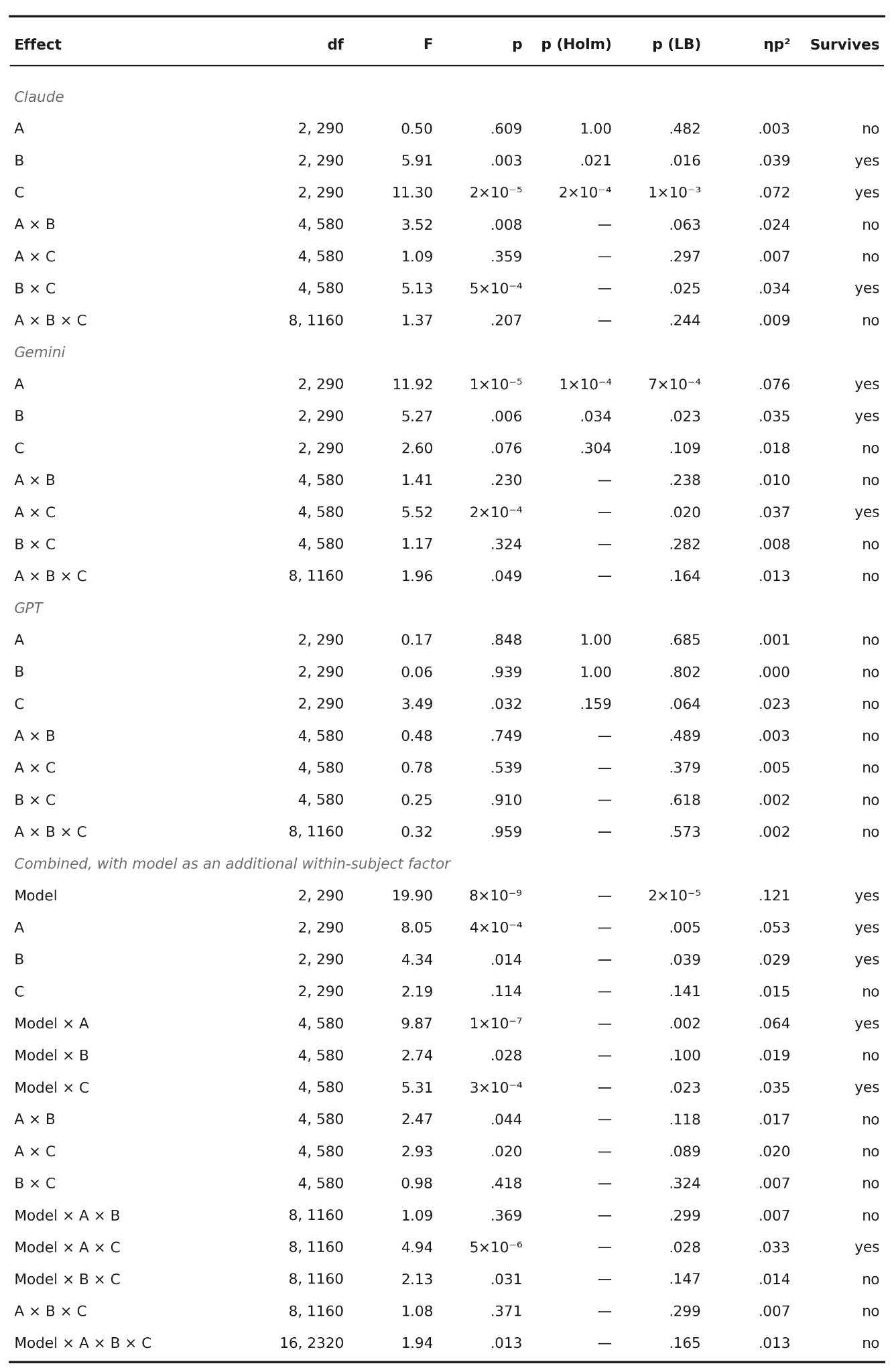}
\end{table}

\begin{table}[tbp]
\centering
\caption{Planned contrasts for rubric and calibration effects on absolute and signed error. Paired t-tests on per-assignment error (df = 145), with the Holm correction applied across the full family of fifteen contrasts. Wilcoxon signed-rank tests are reported uncorrected as a distribution-free check. Δ is the mean difference in marks between the first- and second-named conditions, so negative values on \textbar error\textbar{} indicate lower absolute error under the first-named condition and signed contrasts indicate shifts in the mark level. dz, Cohen's dz for paired designs.}\label{tab:S3}
\includegraphics[width=\linewidth]{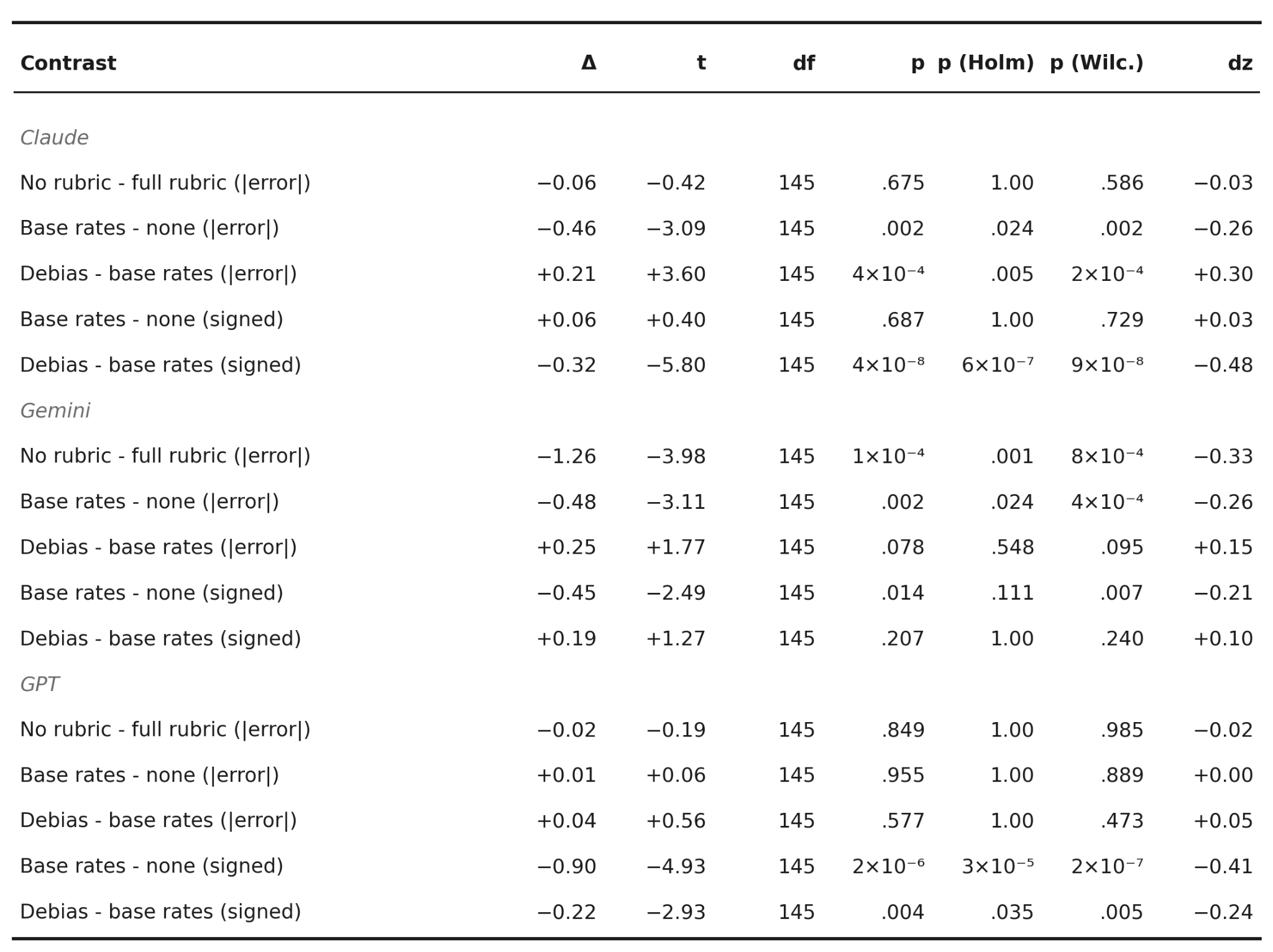}
\end{table}

\begin{table}[tbp]
\centering
\caption{Counterfactual per-institution prompt selection and its transfer. For each model and institution: the locally optimal configuration on that institution's calibration subset (n assignments as shown), its local RMSE, and the rank the deployed pooled optimum would attain among the 27 cells locally. ρ columns give Spearman correlations between institutions' full cell orderings. Per-institution optima were never deployed; all reported test-set results use the pooled configurations of Supplementary Table S1.}\label{tab:S4}
\includegraphics[width=\linewidth]{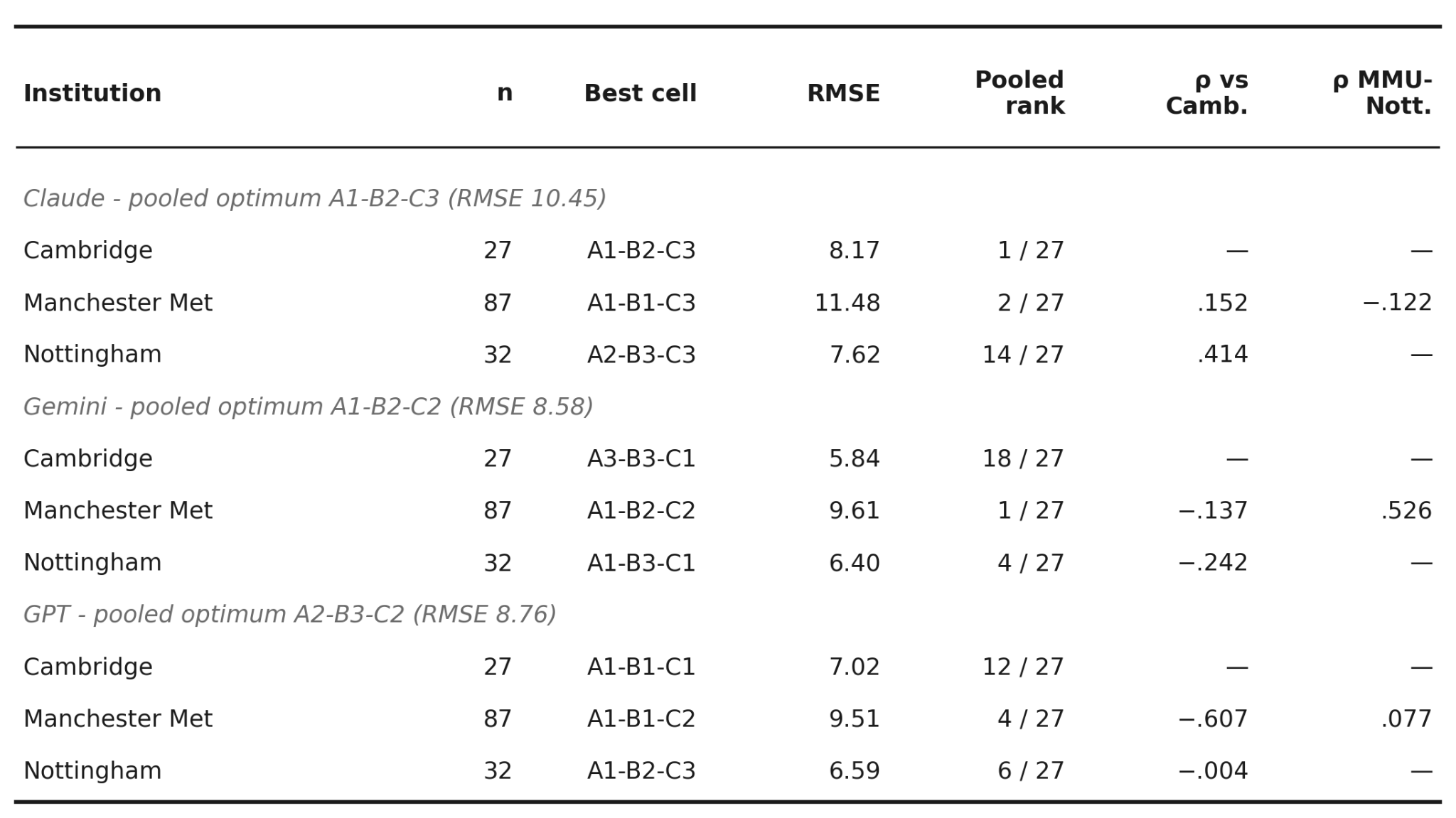}
\end{table}

\section*{Supplementary Note S3: Aggregation strategies and pooled accuracy}

Supplementary Table S5 reports pooled test-set accuracy (n = 608) for the three individual models and four aggregation strategies: the per-assignment median, the unweighted mean, the surprisingly popular answer, and the inverse-RMSE weighted mean used as the headline ensemble. Aggregation delivered modest and non-dominant gains. The weighted mean attains the best MAE, RMSE and QWK, and the median the best exact agreement, but the strongest single model (Claude) retains the highest rank correlation and normalised QWK; no method dominates across columns, consistent with the shared error structure reported in the main text.

Two features of the table warrant a note. First, exact agreement is not comparable across rows: mean-based aggregates produce non-integer marks that rarely coincide exactly with a human mark, which accounts for their collapsed EA values (5.6--5.9\%) despite otherwise competitive accuracy. Second, mean bias is negative for every method (−1.5 to −3.6 marks), so aggregation moderates but does not remove the systematic under-marking documented in the main text.

\begin{table}[tbp]
\centering
\caption{Pooled test-set accuracy for individual models and aggregation strategies. Held-out test set, all institutions pooled (n = 608). † indicates an aggregation over the three models' marks. ρ, Spearman correlation with the human mark; EA/AA/BA, exact, adjacent, and band agreement; QWK and nQWK as defined in Section 3.5; MAE and RMSE in marks; Bias, mean signed error (machine − human). Bold marks the best value in each column. EA values for mean-based aggregates are not comparable to integer-valued methods, as averaging yields non-integer marks.}\label{tab:S5}
\includegraphics[width=\linewidth]{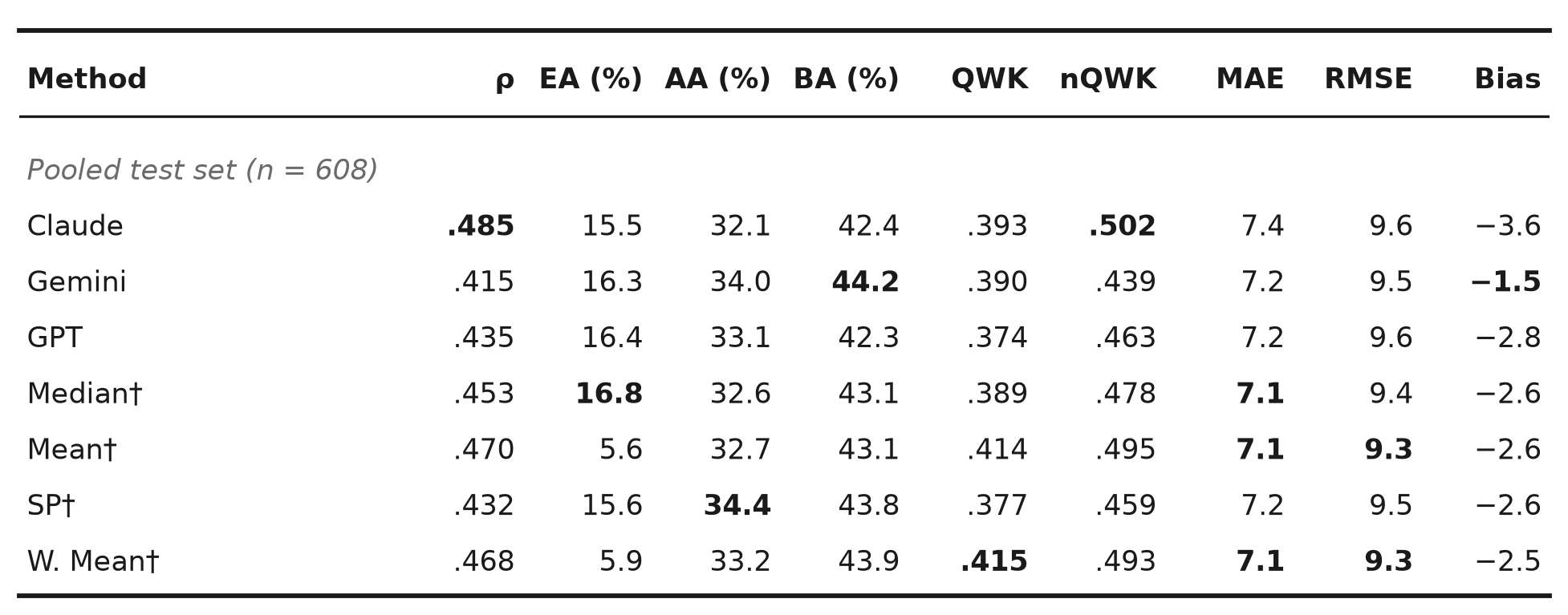}
\end{table}

\section*{Supplementary Note S4: Reliability subset composition and test--retest stability}

The reliability subset comprised 165 assignments drawn to cover the mark distribution at each institution: 46 from Cambridge (15 First, 15 Upper Second, 11 Lower Second, 5 Third), 70 from Manchester Met (15, 14, 16, 13, and 12 Fail), and 49 from Nottingham (14, 16, 15, 4). Failing scripts occur only at Manchester Met, mirroring the institutional mark distributions described in Section 3.1, so the Fail band's stability estimates rest on a single institution.

Supplementary Table S6 reports the two reliability quantities. Within model, marking is close to deterministic: across five repeated runs, intraclass correlations are .998 for Claude, .999 for Gemini, and .968 for GPT, with within-assignment standard deviations of 0.05, 0.01 and 0.73 marks and band-stable percentages of 98.2, 99.4 and 86.7 respectively -- GPT is the only model whose repeated marks cross band boundaries with any regularity. Between models, concordance on the test set is materially lower (ICC(2,k) = .847--.920; α = .638--.791), and since ICC(2,k) indexes the reliability of the three-model average rather than of any single marker, the single-model equivalents are lower still (ICC(2,1) = .648 at Nottingham, .691 at Cambridge and .793 at Manchester Met; .776 pooled). The contrast between near-perfect within-model stability and moderate between-model concordance indicates that disagreement among the models is systematic rather than stochastic: rescoring reproduces each model's marks almost exactly, including their divergences from one another. The independence expectation reported in the main text (18.8\%) is the sum of the products of the three models' marginal below-human and above-human proportions on the test set. It falls below 25\% because exact matches are retained as a third outcome rather than being redistributed across the two error directions.

\begin{table}[tbp]
\centering
\caption{Within-model test--retest stability and inter-model concordance. Upper panel: five repeated scoring runs on the reliability subset (n = 165 assignments) under the deployed configurations; ICC, intraclass correlation across runs, ICC(2,1), two-way random effects, single measures, absolute agreement; sd and Range, mean within-assignment standard deviation and range of marks; Band-stable \%, share of assignments receiving the same degree classification on every run. Lower panel: inter-model concordance on the held-out test set, reporting ICC(2,k) for the three-model average and Krippendorff's α.}\label{tab:S6}
\includegraphics[width=\linewidth]{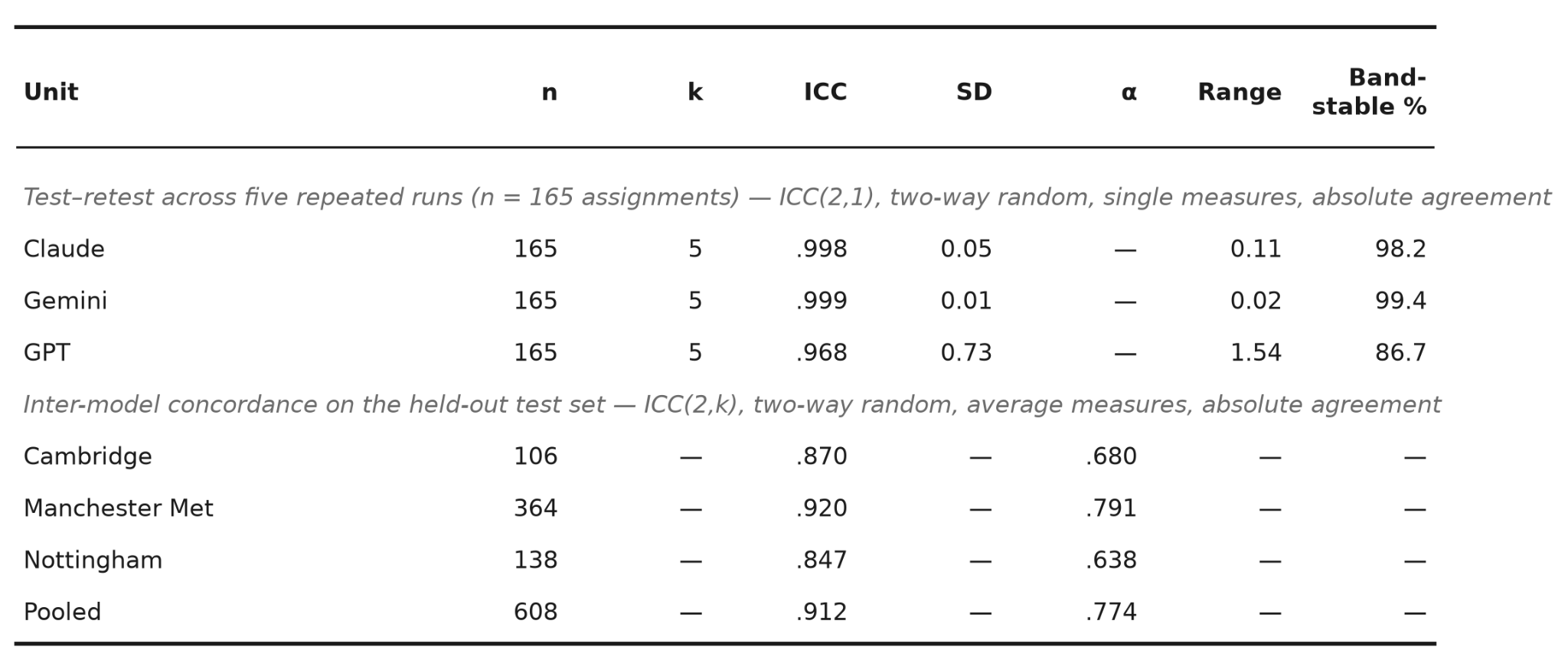}
\end{table}

A system that returns the same mark every time, however imperfect, provides what human moderation has always lacked, i.e., a fixed reference against which drift in human marking standards, between markers, cohorts or years, becomes measurable. Used descriptively rather than authoritatively, machine marks could make the (in)consistency of human marking visible without ever determining a student's grade.

\section*{Supplementary Note S5: Self-reported confidence from machine marking}

Each scoring call elicited a self-reported confidence rating (0--100) alongside the mark. The ratings were collected for completeness but excluded from all aggregation and analysis after calibration-split checks showed them to be uninformative. As Supplementary Table S7 indicates, each model emitted a near-constant, provider-specific value: standard deviations of 3--5.4 points around means of 60.1 (Claude), 87.9 (Gemini) and 76.5 (GPT), with a single modal value accounting for 39--56\% of all calls. Correlations between confidence and absolute error were never in the informative direction -- Claude's was null (r = −.001), while Gemini's and GPT's were small but significantly positive (r = .049 and .113), higher confidence accompanying larger error. Inter-model weighting in the ensembles therefore relies on demonstrated calibration accuracy rather than self-assessment.

\begin{table}[tbp]
\centering
\caption{Self-reported confidence is near-constant within model and uninformative about error. Distribution of self-reported confidence (0--100) across all calibration-split scoring calls (153 assignments × 27 configurations), by model, with the Pearson correlation between confidence and per-assignment absolute error and its p value. Mode (share) gives the most frequent value and the proportion of calls emitting it.}\label{tab:S7}
\includegraphics[width=\linewidth]{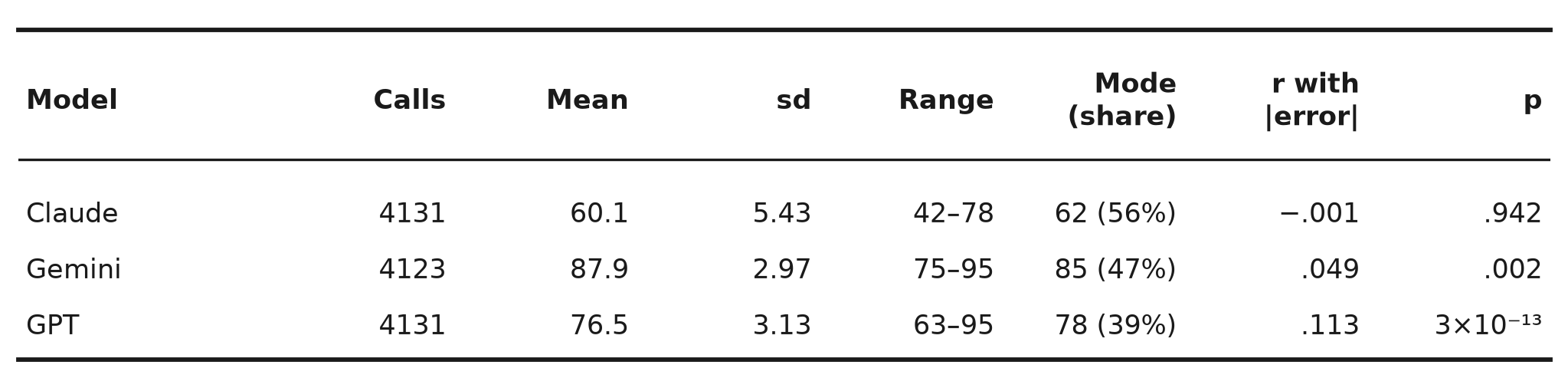}
\end{table}

\section*{Supplementary Note S6: Assessment format and scoring error}

Assessment format was derived from the wording of each assignment brief, there being no format field in the institutional metadata; the analysis is exploratory throughout. Analyses use the held-out test split (complete cases, weighted-mean ensemble) and the four formats with ten or more submissions, three further submissions across two rarer formats being excluded. Within Manchester Met, ensemble MAE ranged from 7.7 to 9.1 across the four formats (Kruskal--Wallis on absolute error, p = .41; format term p = .81 in a model adjusting for the human mark and its square), and within Nottingham from 4.9 to 5.7 (p = .62; adjusted p = .56). Signed error behaves differently. At Manchester Met the format term is reliable (Kruskal--Wallis on signed error, H(3) = 13.87, p = .003; partial F(3, 355) = 23.13, p = 1.1 × 10⁻¹³ under the same adjustment), with the ensemble under-marking research and lab reports by 6.0 marks on average against 1.4 marks for coursework essays. The essay--report contrast is post hoc, but is the only one of the six pairwise comparisons to survive Holm correction (t = 3.77, corrected p = .0013). The report effect holds for each model individually (5.4--6.6 marks); the essay baseline does not, Gemini being essentially unbiased on coursework essays (+0.02), so the 1.4-mark figure is a property of the ensemble rather than a shared one. Cambridge comprises examination essays only, so no within-institution contrast is possible there, and the examination--coursework comparison across institutions is confounded with institution. On the available within-institution evidence, assessment format does not account for the differences in absolute scoring error between institutions, though within Manchester Met it modulates the extent of under-marking.

\section*{Supplementary Note S7: Linguistic feature associations and per-model regressions}

Supplementary Table S8 defines the 33 features extracted from each assignment: 32 linguistic features spanning five domains established in the automated writing assessment literature -- lexical diversity and sophistication, readability, syntactic and structural complexity, discourse and cohesion, and writing error -- together with word count. Extraction methods and the libraries used are given alongside each domain.

\begin{table}[tbp]
\centering
\caption{Definitions and extraction methods for the 33 assignment-level features. The 32 linguistic features span five domains established in the automated writing assessment literature -- lexical diversity and sophistication, readability, syntactic and structural complexity, discourse and cohesion, and writing error -- and are analysed together with word count. Extraction used spaCy and NLTK preprocessing in combination with domain-specific libraries: \emph{lexicalrichness} for diversity indices, \emph{wordfreq} for lexical sophistication, \emph{textstat} for readability formulae, and \emph{symspellpy} for grammar, spelling and punctuation detection (Section 3.6).}\label{tab:S8}
\includegraphics[width=\linewidth]{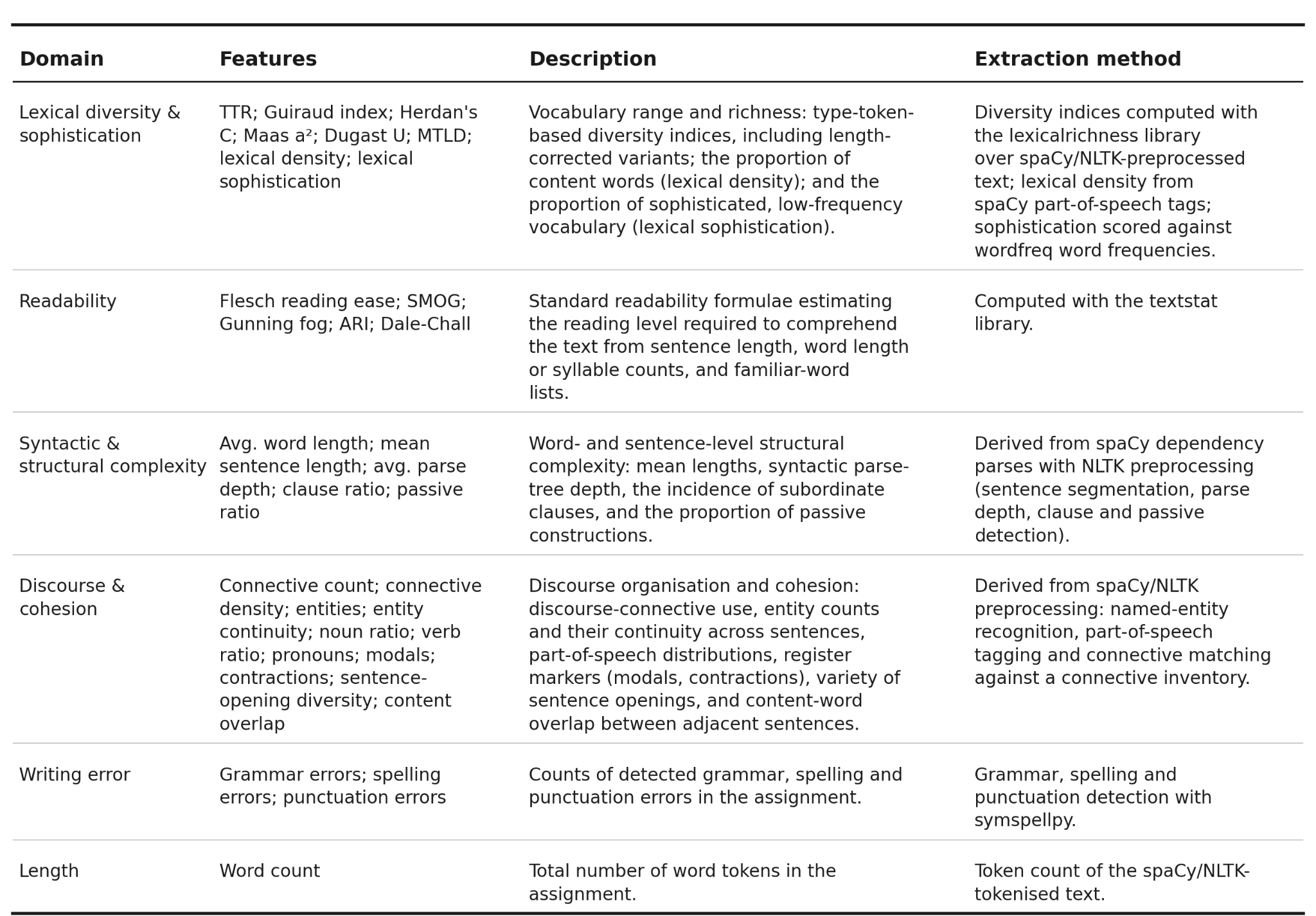}
\end{table}

Supplementary Fig. S1 reports the full matrix of Holm-corrected Spearman correlations among the five mark variables (the human mark, the three individual model marks, and the weighted-mean ensemble) and the 33 features on the held-out test set (n = 608). Two properties of the matrix are consequential for the analyses in the main text. First, human marks correlate reliably with only five features (lexical sophistication, Flesch reading ease, Dale--Chall, grammar errors and spelling errors), whereas the weighted-mean ensemble correlates reliably with 20 of the 33 -- in the same directions, but at consistently greater magnitude for 28 of the 33 features -- with each individual model showing a similarly broadened profile. Second, the features themselves intercorrelate substantially, particularly within domain, indicating that several measures capture overlapping constructs rather than distinct properties of writing; this motivates the dimension reduction reported in the main text.

\begin{figure}[tbp]
\centering
\includegraphics[width=\linewidth]{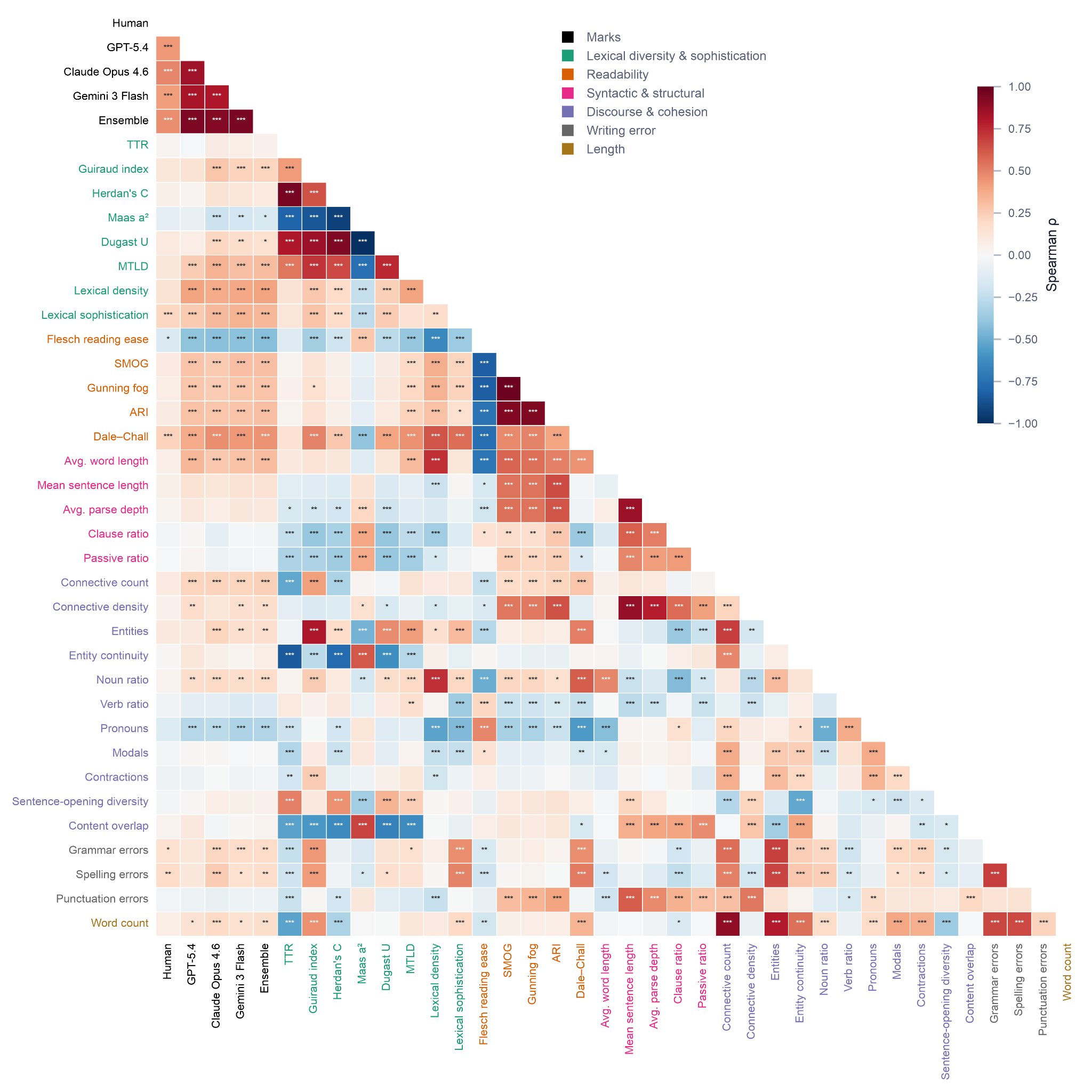}
\caption{Machine marks correlate with a far broader set of surface linguistic features than human marks. Spearman rank-order correlations among the five mark variables (Human; GPT-5.4; Claude Opus 4.6; Gemini 3 Flash; weighted-mean ensemble) and the 33 extracted features, held-out test set (n = 608), with rows and columns grouped and colour-coded by linguistic domain. Asterisks denote significance after Holm correction across all 703 unique pairwise comparisons (*p \textless{} .05, **p \textless{} .01, ***p \textless{} .001). Human marks correlate reliably with 5 features; the weighted-mean ensemble with 20 of the 33, each in the same direction as the corresponding human correlation. Across all 33 features, the ensemble coefficient is the larger in magnitude for 28. The substantial within-domain intercorrelation among features motivates the principal component analysis reported in the main text.}\label{fig:S1}
\end{figure}

Supplementary Fig. S2 gives the loadings and interpretations of the six components retained from the principal component analysis of the standardised features, which jointly explain 79.8\% of feature variance.

\begin{figure}[tbp]
\centering
\includegraphics[width=\linewidth]{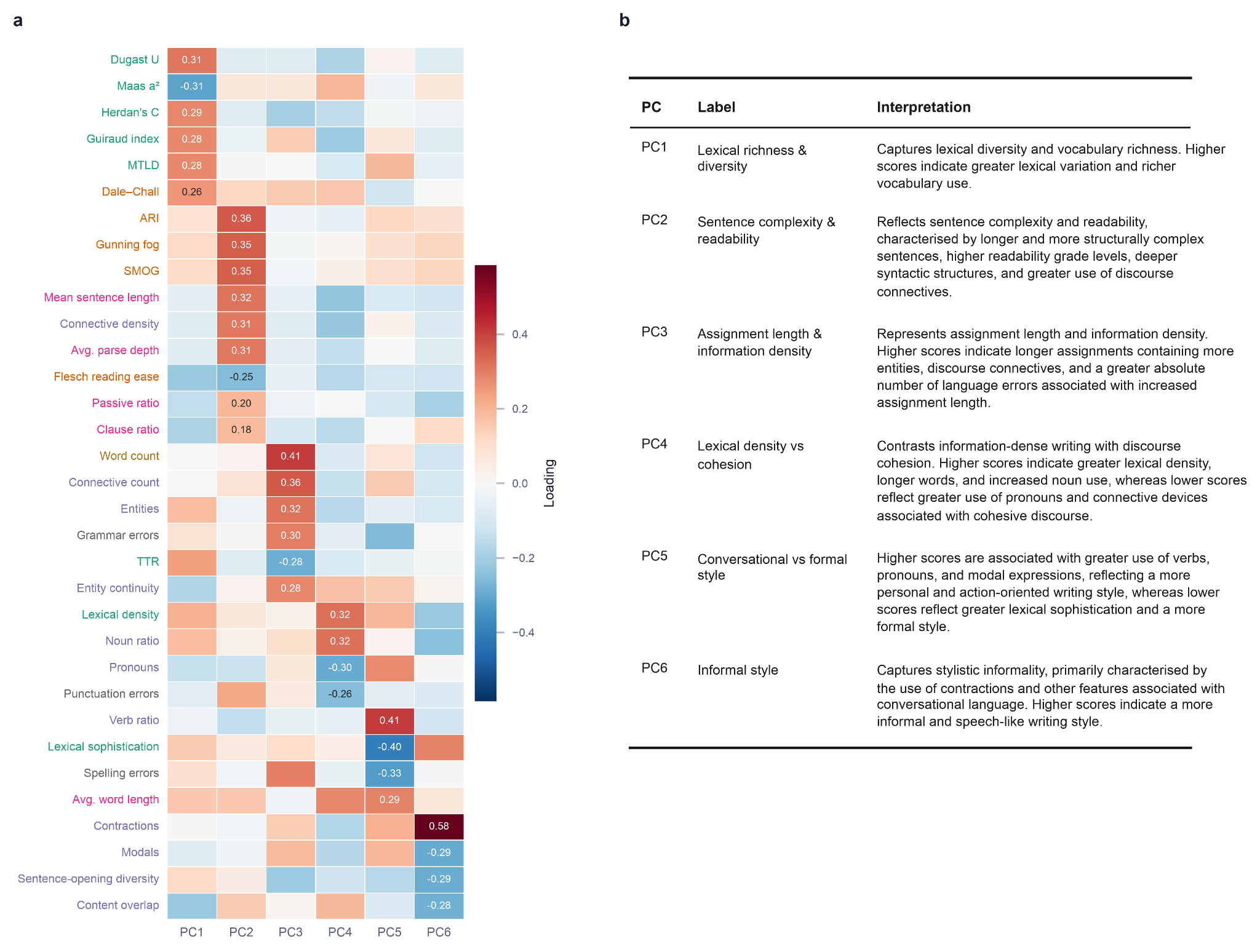}
\caption{Loadings and interpretation of the six retained principal components. (a) Loadings of the 33 standardised features (rows, colour-coded by linguistic domain) on the six retained components; the strongest loadings are annotated. (b) Label and substantive interpretation of each component. The six components jointly explain 79.8\% of feature variance (24.1\%, 19.2\%, 16.2\%, 10.7\%, 6.8\% and 2.8\% respectively); retention was determined by the elbow criterion on the scree plot (Section 3.6).}\label{fig:S2}
\end{figure}

Supplementary Table S9 reports the regressions of Table 2 fitted separately for each individual model's marks, together with the institutional covariate coefficients omitted from Table 2 for space. The asymmetry documented in the main text is not an artefact of ensembling: each individual model's marks are substantially more predictable from the six linguistic components and institutional covariates together than human marks are (R² = 0.320 for GPT, 0.365 for Gemini and 0.328 for Claude, against 0.111 for human marks), and sentence complexity and readability (PC2) is a significant predictor of every model's marks while remaining unrelated to the human mark. Ensembling sharpens the shared conditioning (ensemble R² = 0.385) but does not create it. The institutional covariates in every model recover the ordering of mean marks described in Section 3.1, with Cambridge highest, and are included so that the linguistic coefficients reflect within-institution variation rather than between-institution differences in mark distributions.

\begin{table}[tbp]
\centering
\caption{Per-model regressions of machine marks on the six linguistic components, and institutional covariates from the human and ensemble models. Upper panel: linear regressions of each individual model's marks on the six PCA-derived linguistic components, held-out test set (n = 608), paralleling the human and ensemble models of Table 2; institution was included as a dummy-coded covariate (Cambridge as reference). Unstandardised coefficients (b) with standard errors and two-sided p values; boldface indicates p \textless{} .05; model fit gives R² and the F statistic. Lower panel: institutional covariate coefficients from the human and ensemble models of Table 2, omitted there for space. PC2 (sentence complexity \& readability) significantly predicts every model's marks, and per-model R² (0.320--0.365) substantially exceeds that of the human-mark model (0.111).}\label{tab:S9}
\includegraphics[width=\linewidth]{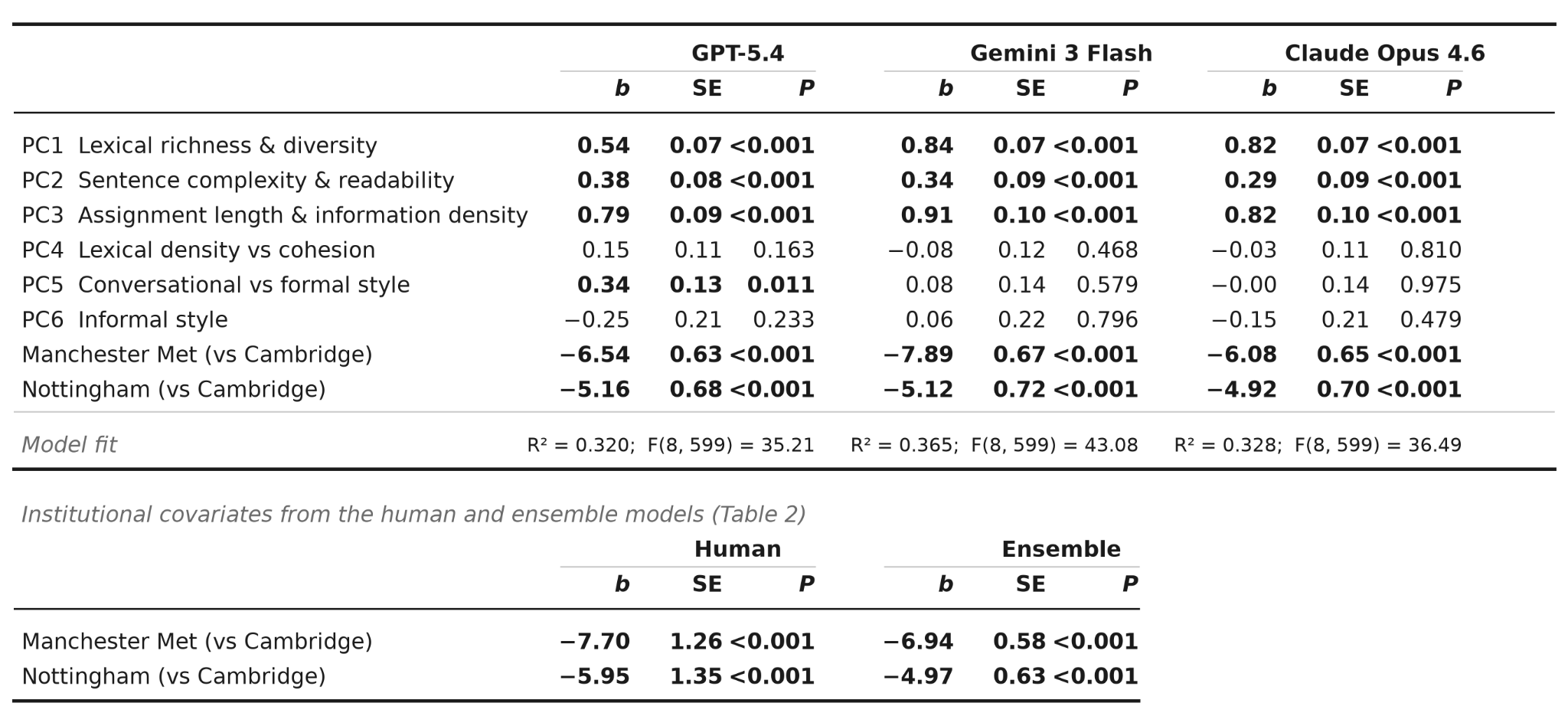}
\end{table}

\end{document}